\documentclass[final,3p,times,twocolumn]{elsarticle}
\usepackage{amssymb}
\usepackage{amsmath}
\usepackage{amsthm}

\usepackage[ruled,linesnumbered]{algorithm2e}
\usepackage{color, xcolor} % 颜色包，color 必须导入，xcolor 建议导入
\usepackage{subfigure} %插入多图时用子图显示的宏包
\usepackage{tabularx}
\usepackage{array}
\usepackage{multirow}
\usepackage{graphicx}
\usepackage{ragged2e}
\usepackage{bm}
\usepackage{bbding}
\usepackage{soul}
\usepackage{longtable}
\usepackage{booktabs}  % For \toprule, \midrule, \bottomrule
\usepackage{siunitx}   % For S columns (numerical alignment)
\usepackage{makecell}
\usepackage[numbers]{natbib}
\usepackage{pgfplots}
\usepackage{listings}         % Required for code formatting inside tcolorbox
\usepackage[most]{tcolorbox}  % Required for creating the colored boxes
\usepackage{enumitem} % 引入新宏包，用于定制列表
\usepackage{url}
\usepackage{hyperref}

\lstdefinestyle{simplecode}{
    basicstyle=\small\ttfamily,  % 使用小号打字机字体
    breaklines=true,             % 关键：允许自动换行
    breakatwhitespace=true,      % 仅在空格处换行，样式更美观
    frame=none,                  % 无边框，与背景融为一体
    showstringspaces=false,      % 不显示字符串中的空格标记
}

\newtcolorbox{promptbox}[1]{ % <--- 这里加上 [1]
    enhanced,
    title=\textbf{#1}, % <--- 这里把具体的标题换成 #1
    colback=gray!5!white,
    colframe=gray!60!black,
    fonttitle=\bfseries,
    attach boxed title to top left={yshift=-2mm, xshift=3mm},
    boxed title style={
        colback=gray!60!black,
        arc=1.5mm,
    },
    arc=1.5mm,
    boxrule=1.5pt,
    breakable,
    width=0.49\textwidth, 
}

\newtcbtheorem[auto counter,
    list type=figure,
]{promptFig}{Figure}{
    enhanced,
    colback=gray!5!white,
    colframe=gray!60!black,
    fonttitle=\bfseries,
    attach boxed title to top left={yshift=-2mm, xshift=3mm},
    boxed title style={colback=gray!60!black, arc=1.5mm},
    arc=1.5mm,
    boxrule=1.5pt,
    breakable,
    label separator={:},
    width=0.49\textwidth,
}{fig} % 设置标签前缀，例如 fig:prompt4

\journal{Knowledge-Based Systems}

\begin{document}

\begin{frontmatter}

\title{Knots: A Large-Scale Multi-Agent Enhanced Expert-Annotated Dataset and LLM Prompt Optimization for NOTAM Semantic Parsing}

\author[a]{Maoqi Liu}
\author[a]{Quan Fang\corref{cor1}}
\cortext[cor1]{Corresponding author: qfang@bupt.edu.cn}
\author[b,c]{Yang Yang}
\author[d]{Can Zhao}
\author[b,c]{Kaiquan Cai}
%{bauthor@xxx.com}

\address[a]{organization={Beijing University of Posts and Telecommunications},%Department and Organization
            addressline={Xitucheng Road 10, Haidian District, Beijing, PRC}, 
            city={Beijing},
            postcode={100876}, 
            state={Beijing},
            country={China}}

\address[b]{organization={Beihang University},%
           department={School of Electronic and Information Engineering},
           city={Beijing},
           postcode={100191},
           country={China}}
           
\address[c]{organization={State Key Laboratory of CNS/ATM},%
           city={Beijing},
           postcode={100191},
           country={China}}

\address[d]{organization={Aviation Data Communication Corporation},
           city={Beijing},
           postcode={100191},
           country={China}}

%% Abstract
\begin{abstract}
Notice to Air Missions (NOTAMs) serve as a critical channel for disseminating key flight safety information, yet their complex linguistic structures and implicit reasoning pose significant challenges for automated parsing. Existing research mainly targets surface-level tasks like classification and named entity recognition, lacking deep semantic understanding. To address this, we propose \textbf{NOTAM semantic parsing}, a task emphasizing semantic inference and integration of aviation domain knowledge for structured, inference-rich outputs. To support this task, we construct \textit{Knots} (\textbf{K}nowledge and \textbf{NOT}AM \textbf{S}emantics), a high-quality dataset of 12,347 expert-annotated NOTAMs covering 194 Flight Information Regions, enhanced through a multi-agent collaborative framework for comprehensive field discovery. We systematically evaluate various prompt engineering strategies and model adaptation techniques, achieving significant improvements in aviation text understanding and processing. Our experimental results demonstrate the effectiveness of the proposed approach and provide valuable insights for automated NOTAM analysis systems. Our code is available at \url{https://github.com/Estrellajer/Knots}.
\end{abstract}

%% Keywords

\end{frontmatter}

\section{Introduction}
\label{sec:intro}

\begin{figure*}[t]
\centering
\includegraphics[width=0.98\textwidth]{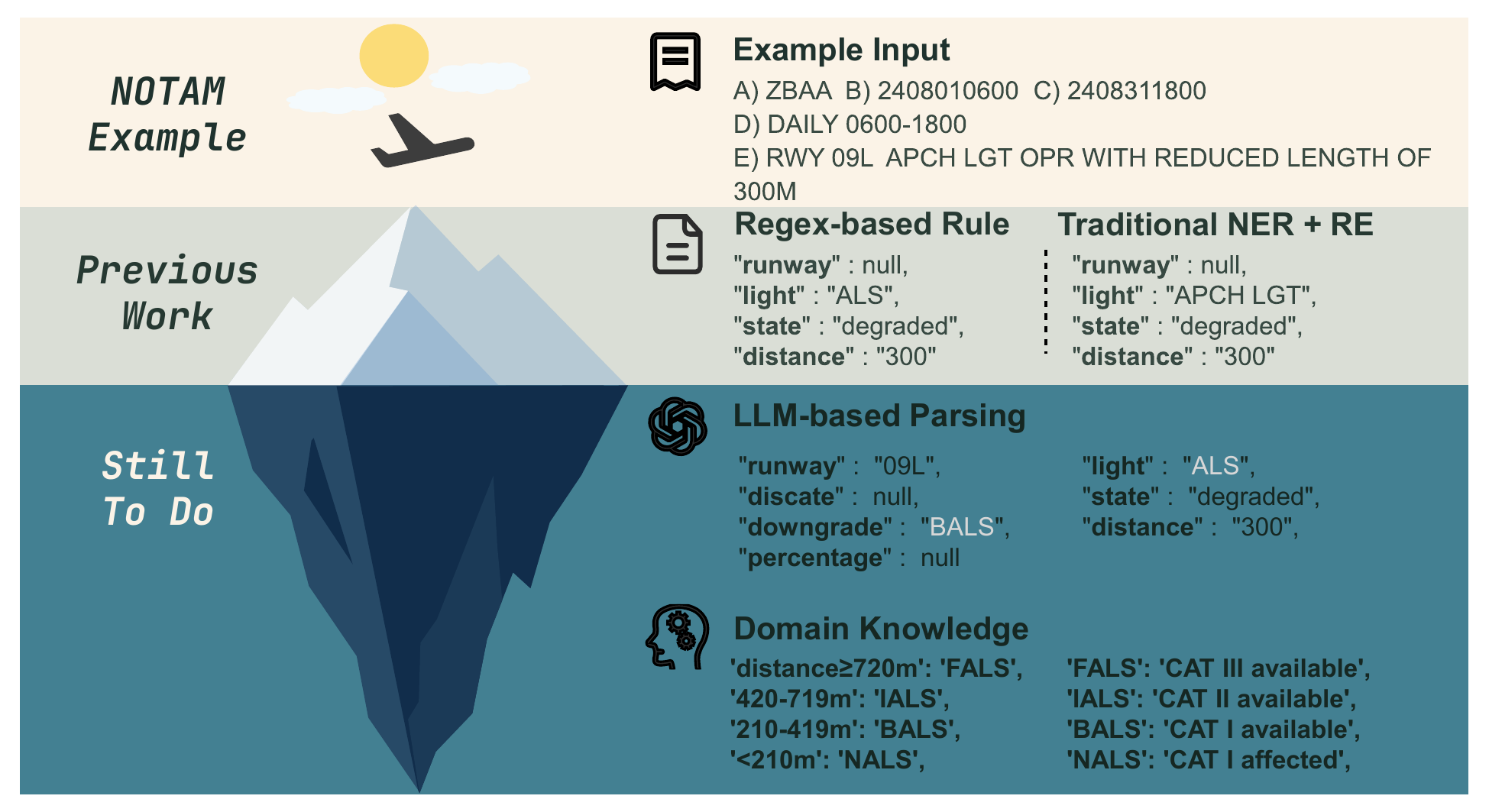}
\caption{An illustrative comparison of different paradigms for NOTAM information extraction. The tip of the iceberg represents traditional methods like regex-based rules and NER, which only scratch the surface by extracting explicitly stated keywords. In contrast, the submerged part visualizes the depth required by the "NOTAM parsing" task. This deeper analysis involves semantic understanding and inference to produce a highly structured, hierarchical, and application-ready output, a challenge well-suited for modern LLMs.}
\label{fig:teaser}
\end{figure*}

Notice to Air Missions (NOTAMs) serve as the core channel for disseminating time-sensitive and critical information regarding airspace restrictions, facility outages, and other factors affecting flight planning and safety decisions. With over one million NOTAMs issued globally each year, accurate and timely interpretation of this information is fundamental for maintaining aviation safety standards and operational efficiency \citep{arnold2022knowledge}. Despite their importance, automated processing of NOTAMs faces significant challenges due to their complex linguistic structures, extensive use of domain-specific abbreviations, and the reasoning skills required to extract actionable information.

Existing research on NOTAM analysis has primarily focused on classification, Named Entity Recognition (NER), and basic content filtering \citep{bravin2020automated, clarke2021natural}. While these approaches are valuable for organizing and categorizing NOTAM content, they remain essentially surface-level, identifying only explicit textual elements and lacking the deep semantic understanding necessary for practical operational use. This limitation is especially evident when NOTAMs include implicit information, abbreviated references, or complex spatiotemporal relationships requiring domain expertise for correct interpretation.

To address this gap, we hereby propose \textbf{NOTAM parsing}—a novel task that extends beyond traditional information extraction. This task involves generating comprehensive structured representations via semantic inference, contextual analysis, and incorporation of aviation domain knowledge. As illustrated in Figure~\ref{fig:teaser}, unlike conventional methods that assume the target information exists explicitly in the source text, NOTAM parsing demands the ability to infer operational implications, normalize abbreviations, and unveil latent relationships among fields. For instance, while a traditional extraction system might identify ``RWY 09L degraded 300M'', NOTAM parsing infers from domain knowledge that this indicates degradation of the Basic Approach Lighting System (BALS).

Historically, rule-based and template-matching approaches, although effective in certain scenarios, have struggled with the diversity and unstructured nature of NOTAM texts \cite{clarke2021natural}. Consequently, such methods often fail to resolve semantic ambiguities and meet the implicit reasoning demands required, thereby limiting their generalization capabilities and practical utility.

The recent emergence of Large Language Models (LLMs) and their breakthroughs in natural language understanding present promising opportunities for the NOTAM parsing task. With powerful semantic modeling capabilities, LLMs can accurately comprehend complex instructions and contextual information, offering a viable pathway to address the inferential challenges inherent in NOTAMs. Although no prior research has specifically explored the application of LLMs to NOTAM parsing, advances in related areas such as complex instruction following and general-purpose information extraction (e.g., \citep{morarasu2024aidriven}) provide a solid technical foundation for our study.

To bridge this gap, our paper pioneers a comprehensive methodology for the task of NOTAM semantic parsing. Our approach is twofold, addressing the core challenges from both a data-centric and a model-centric perspective. First, recognizing that progress is contingent on high-quality resources, we establish a new benchmark by detailing the principles and expert-led process for creating a dataset geared for deep semantic inference. Second, we shift focus to the practical application of LLMs, presenting a systematic investigation into how these powerful models can be effectively prompted and optimized to handle the unique linguistic and safety-critical nature of the aviation domain.

The primary contributions of this paper are as follows:
\begin{enumerate}
    \item We construct a meticulously annotated dataset for NOTAM parsing covering key operational fields, ensuring both data quality and practical relevance.
    \item We design and implement a multi-agent collaborative pipeline for automated discovery of new fields, which can support more comprehensive future evaluations of NOTAM parsing systems.
    \item We systematically evaluate various prompt engineering strategies for the NOTAM parsing task and propose customized improvements informed by aviation domain knowledge.
\end{enumerate}

The remainder of this paper is organized as follows. Section 2 reviews related work in aviation NLP, information extraction, and multi-agent systems. Section 3 establishes the formal problem definitions and notation. Section 4 introduces our dataset construction methodology. Section 5 describes the design of our multi-agent framework. Section 6 presents our comprehensive NOTAM parsing methodology, including prompt engineering strategies and domain-specific optimization techniques. Section 7 provides a comprehensive experimental evaluation. Finally, Section 8 summarizes the research findings and discusses their implications and future research directions.

\section{Related Work} 
\label{sec:related_work} 

\subsection{NLP Applications in Aviation Domain} 

Natural Language Processing has become transformative for aviation safety, with NOTAM analysis being particularly challenging due to its safety-critical nature \citep{mogillo-dettwilerfiltering, mi2022notam}. 

\textbf{NOTAM Processing}: Early work established traditional NLP foundations using transformer architectures for filtering and inconsistency detection \citep{bravin2020automated}. Comprehensive workflows integrated TF-IDF, topic modeling, and NER for automated segmentation \citep{clarke2021natural}. BERT models trained on 1.2 million NOTAMs achieved significant scalability improvements \citep{arnold2022knowledge}. However, challenges persist: ambiguous abbreviations, semantic-practical mismatches, and regional variations compromise safety \citep{morarasu2024aidriven}.

\textbf{Broader Applications}: Contemporary research expanded to flight operations management. Spoken instructions were integrated into trajectory prediction \citep{guo2024integrating}, graph-based approaches modeled trajectory relationships \citep{fan2024global}, and trajectory prediction was framed as language modeling \citep{luo2025large}. Specialized applications include pilot phraseology assessment \citep{liu2024nlp} and flight phase classification \citep{nanyonga2023aviation}. These advances reveal the need for adaptive approaches handling aviation's complex, safety-critical nature.

\subsection{Large Language Models for Information Extraction} 

LLMs have revolutionized information extraction, transitioning from discriminative to unified generative frameworks \citep{xu2024survey, brown2020languagemodelsfewshotlearners}. 

\textbf{Prompt Engineering}: Core techniques include zero-shot and few-shot prompting \citep{Brown2020, Sahoo2025}, and Chain-of-Thought (CoT) enabling stepwise reasoning \citep{Wei2022}. Advanced variants include Automatic CoT \citep{Zhang2022}, Self-Consistency \citep{Wang2022}, and Logical CoT \citep{zhao2023enhancing}. Structural frameworks like Tree-of-Thoughts \citep{Yao2023a} and Graph-of-Thoughts \citep{Yao2023b} enable non-linear reasoning. Retrieval-Augmented Generation \citep{Lewis2020} and Chain-of-Verification \citep{Dhuliawala2023} address factual accuracy. Challenges remain in robustness and domain adaptation \citep{Yuan2024}.

\textbf{Unified Extraction}: Contemporary LLMs unify named entity recognition, relation extraction, and event extraction \citep{xu2024survey}. Frameworks like UIE, InstructUIE, and Code4UIE excel in low-resource scenarios through in-context learning \citep{li2023codeie} and instruction-based fine-tuning \citep{wang2023instructuie}. Code-style prompting \citep{sainz2024gollie} and hierarchical representations \citep{li2024knowcoder} enhance accuracy. Aviation domains present unique challenges requiring sophisticated adaptation due to dynamic semantics and safety requirements.

\subsection{Multi-Agent Systems for Complex Task Decomposition} 

LLM-based multi-agent systems establish new paradigms for collaborative problem-solving through distributed intelligence \citep{Tran2025MultiAgentCM}. 

\textbf{Collaborative Architectures}: Early frameworks emphasized brain, perception, and action components \citep{xi2023}. Communication topologies were systematically investigated \citep{guo2024multiagent}, with taxonomies categorizing strategies into merging, ensemble, and cooperative approaches \citep{lu2024}.

\textbf{Advanced Mechanisms}: Sophisticated approaches leverage cognitive psychology principles. Planning architectures emphasize specialization and hierarchical decomposition \citep{han2024}. Debate-driven divergent thinking enables robust exploration \citep{liang2023}. Social psychology-inspired strategies \citep{zhang2023} and Theory-of-Mind collaboration \citep{li2023a} provide cognitive foundations for interaction.

\textbf{Implementation Frameworks}: Modular frameworks like CAMEL \citep{li2023b} and AutoGen \citep{wu2024} facilitate systematic design. Domain-specific implementations in digital twins \citep{he2025} and software engineering \citep{lambiase2024} demonstrate versatility. Despite this potential, application to safety-critical aviation domains requiring expert knowledge integration remains unexplored, motivating our approach.

\section{Preliminaries}
\label{sec:preliminaries}

We establish formal definitions for the NOTAM parsing problem, encompassing both foundational field extraction and emergent field discovery tasks. Our framework addresses two complementary challenges: extracting predefined operational fields with high accuracy, and discovering novel, contextually relevant fields that emerge from diverse NOTAM expressions.

\textbf{Notation and Problem Space:} Let $\mathcal{X}$ denote the space of NOTAM natural language texts, and $\mathcal{Y}$ represent the valid structured representation space for field extractions. We define $\mathbb{F} = \{f_1, \ldots, f_K\}$ as the predefined foundational field set with $K$ critical operational fields (e.g., runway status, effective periods, restrictions), while $\mathcal{U}$ represents the potential emergent field space where $|\mathcal{U}| \gg |\mathbb{F}|$. Our annotated dataset is denoted as $\mathbb{D} = \{(x^n, y^n)\}_{n=1}^{N}$.

\textbf{Task 1 - Foundational Field Extraction:} The primary extraction task maps NOTAM text to structured representations through semantic inference rather than simple text matching. We formalize this as $G_\theta : \mathcal{X} \rightarrow \mathcal{Y}$, where the optimal output is determined by $\hat{y} = G_\theta(x) \triangleq \arg\max_{y \in \mathcal{Y}} P_\theta(y|x)$. The model parameters $\theta$ can be optimized through various paradigms including zero-shot prompting (leveraging pre-trained knowledge), in-context learning with domain-specific examples, or supervised fine-tuning on annotated NOTAM data.

\textbf{Task 2 - Emergent Field Discovery:} Beyond predefined fields, NOTAMs often contain operationally significant information that cannot be anticipated in advance, such as aircraft-specific restrictions or unusual weather conditions. For a given NOTAM text $x$, we aim to discover emergent fields $\mathcal{E}(x) = \{e_1, e_2, \ldots, e_M\} \subset \mathcal{U}$, where each emergent field $e_i = (n_i, d_i, v_i, s_i)$ consists of a field name $n_i$, semantic description $d_i$, extracted value $v_i$, and supporting textual evidence $s_i$ from the original NOTAM.

This dual-task formulation enables comprehensive NOTAM understanding: foundational extraction ensures coverage of essential operational information, while emergent discovery provides the potential for more comprehensive NOTAM evaluation in the future.

\begin{figure}[!ht]
\centering
\includegraphics[width=0.48\textwidth]{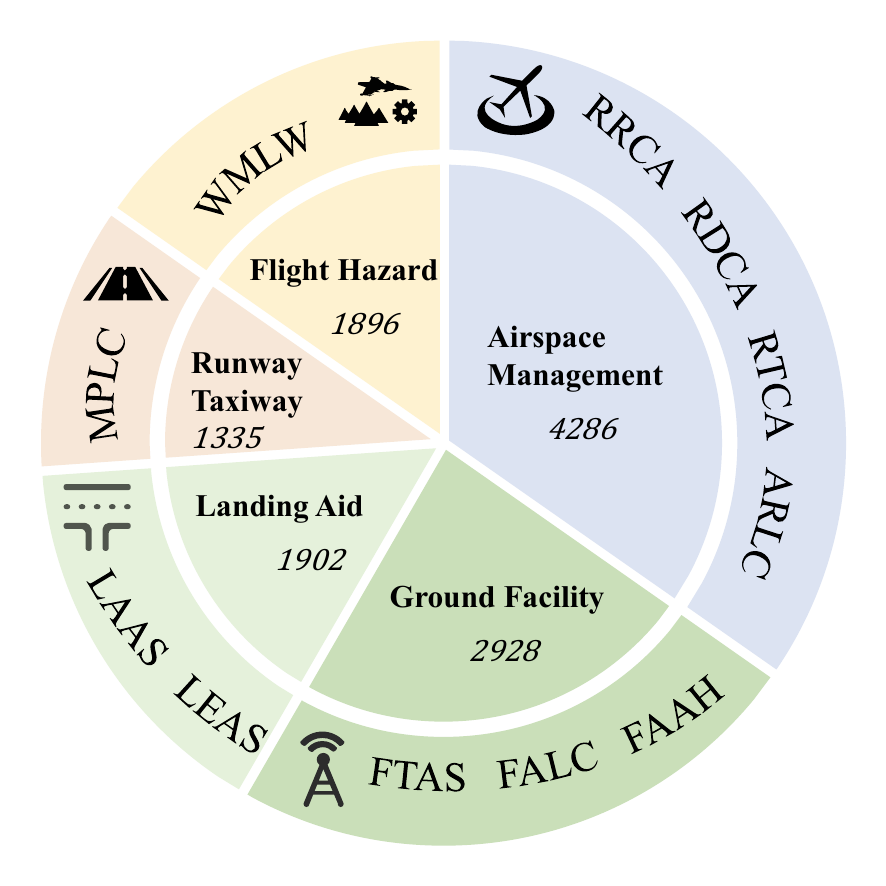}
\caption{Category and subcategory distribution of Q-codes within the NOTAM dataset.}
\label{fig:bing}
\end{figure}

\section{Dataset Construction and Experimental Setup}
\label{sec:dataset}

\subsection{Dataset Overview}

We introduce Knots, a new, large-scale, and comprehensive dataset designed to facilitate research in the automated parsing of Notices to Air Missions (NOTAMs). NOTAMs are critical safety messages that exhibit significant variations in parsing, posing a major challenge for automated systems. As illustrated in Figure~\ref{fig:bing}, Knots is structured around five major operational domains: Airspace Management (\num{4286}), Ground Facility (\num{2928}), Landing Aid (\num{1902}), Runway \& Taxiway (\num{1335}), and Flight Hazard (\num{1896}), encompassing a total of 58 distinct Q-code categories. To capture the diversity of real-world operations, the dataset was collected from global NOTAM broadcasts, comprising \num{12347} valid records from \num{1484} distinct airports, spanning \num{194} Flight Information Regions (FIRs) across seven continents. As detailed in Table~\ref{tab:notam_statistics}, the textual content displays considerable variability, with each NOTAM averaging \num{44.8} words (ranging from \num{3} to \num{1541}) and \num{335.6} characters (ranging from \num{14} to \num{12895}). A primary source of this complexity is the inconsistent adherence to the parsing standards stipulated by the International Civil Aviation Organization (ICAO). This issue is compounded by the dataset's global geographic distribution—sourced predominantly from Europe (\SI{43.99}{\percent}) and Asia (\SI{32.37}{\percent})—which introduces significant regional variations in terminology and formatting conventions. Consequently, Knots serves as a challenging new benchmark for developing robust and geographically-aware automated NOTAM interpretation systems.

\subsection{Data Annotation}
The annotation process was meticulously designed to ensure data quality, domain relevance, and the creation of a benchmark that fosters advanced semantic understanding rather than simple text extraction. The process involved a rigorous, expert-led workflow and resulted in two distinct outputs: the primary annotated dataset and a specialized evaluation set for field discovery.

\begin{table}[!ht]
\centering
\footnotesize \selectfont
\caption{Statistical overview of the NOTAM dataset, detailing Q-code counts, category distribution, and spatial coverage.}
\label{tab:notam_statistics}
\begin{tabular}{lr}
    \toprule
    \textbf{Properties} & \textbf{Value} \\
    \midrule
    \multicolumn{2}{c}{\textbf{\textit{Basic Insight}}} \\
    \midrule
    \textbf{Valid Q-codes} & 12347 \\
    \textbf{Q-code types} & 58 \\
    \textbf{Airports involved} & 1484 \\
    \textbf{FIRs involved} & 194 \\
    \textbf{Continents involved} & 8 \\
    \textbf{Average validity period} & 9.4 days \\
    \textbf{Shortest validity period} & 0.3 days \\
    \textbf{Longest validity period} & 1184 days \\
    \textbf{Median validity period} & 0 days \\
    \textbf{Average word count} & 44.8 \\
    \textbf{Average character count} & 335.6 \\
    \textbf{Average line count} & 13.7 \\
    \midrule
    \multicolumn{2}{c}{\textbf{\textit{Q-Code Category Distribution}}} \\
    \midrule
    \textbf{R} & Airspace Restrictions (3859, 31.25\%) \\
    \textbf{F} & Facilities and Services (2486, 20.13\%) \\
    \textbf{W} & Warning Information (1896, 15.36\%) \\
    \textbf{L} & Lighting Facilities (1424, 11.53\%) \\
    \textbf{M} & Movement Areas (1335, 10.81\%) \\
    \textbf{I} & Instrument Systems (478, 3.87\%) \\
    \textbf{N} & Navigation Facilities (442, 3.58\%) \\
    \textbf{A} & Airspace Organization (232, 1.88\%) \\
    \textbf{P} & Flight Procedures (195, 1.58\%) \\
    \midrule
    \multicolumn{2}{c}{\textbf{\textit{Summary}}} \\
    \midrule
    \textbf{Total NOTAM Records} & 12347 \\
\bottomrule
\end{tabular}
\end{table}

\subsubsection{Annotation Process and Quality Assurance}
Given the vast number of NOTAMs issued annually, we began by randomly sampling from the global NOTAM traffic of the entire year 2024. This approach ensures a manageable data volume while maintaining broad, representative coverage, a characteristic substantiated by the statistics in Table~\ref{tab:notam_statistics}. The initial set of annotation fields was defined by expert aviation dispatchers, who selected foundational fields critical to operational efficiency and safety.

The core of our methodology is a non-extractive, inferential annotation scheme. Unlike traditional information extraction datasets that rely on sequence-labeling formats (e.g., BIO), our annotators were tasked with providing the semantically correct value for each field, irrespective of whether this value could be directly retrieved from the source text. As illustrated in Figure~\ref{fig:teaser}, this often requires annotators to apply domain-specific knowledge to infer, normalize, or calculate values. For instance, a NOTAM might imply a runway closure through technical jargon, and the annotator's role is to explicitly assign the value ``Closed'' to the ``Runway Status'' field. This philosophy ensures the dataset supports the development of models capable of genuine comprehension.

The annotation was performed independently by two expert dispatchers. To ensure consistency and quality, we calculated the Inter-Annotator Agreement (IAA), achieving a Krippendorff's Alpha score of 0.96, which indicates a very high degree of reliability. All discrepancies were subsequently resolved by a third, senior expert to produce the final, gold-standard dataset.

\subsubsection{Evaluation Set for Field Discovery}
A key challenge in NOTAM processing is the presence of low-frequency yet operationally critical information, which is often overlooked by automated systems focused on common fields. Examples include mentions of ``Snow'' (implying runway contamination) or a specific ``Flight Code'' (indicating restrictions on certain aircraft types). While identifying such critical, non-standard fields from scratch is difficult, verifying their importance is a relatively straightforward task for a domain expert.

Leveraging this principle, we created a specialized benchmark for the task of automated field discovery. We randomly sampled 500 NOTAMs from our collection and presented them to an expert. The expert's task was to identify and label any information that was not part of the foundational field set but was nonetheless valuable for safety or efficiency. This process yielded a curated evaluation set that can be used to validate methods for importance-driven information extraction and automated discovery of salient, long-tail data fields.

\begin{figure*}[!ht]
\centering
\includegraphics[width=0.98\textwidth]{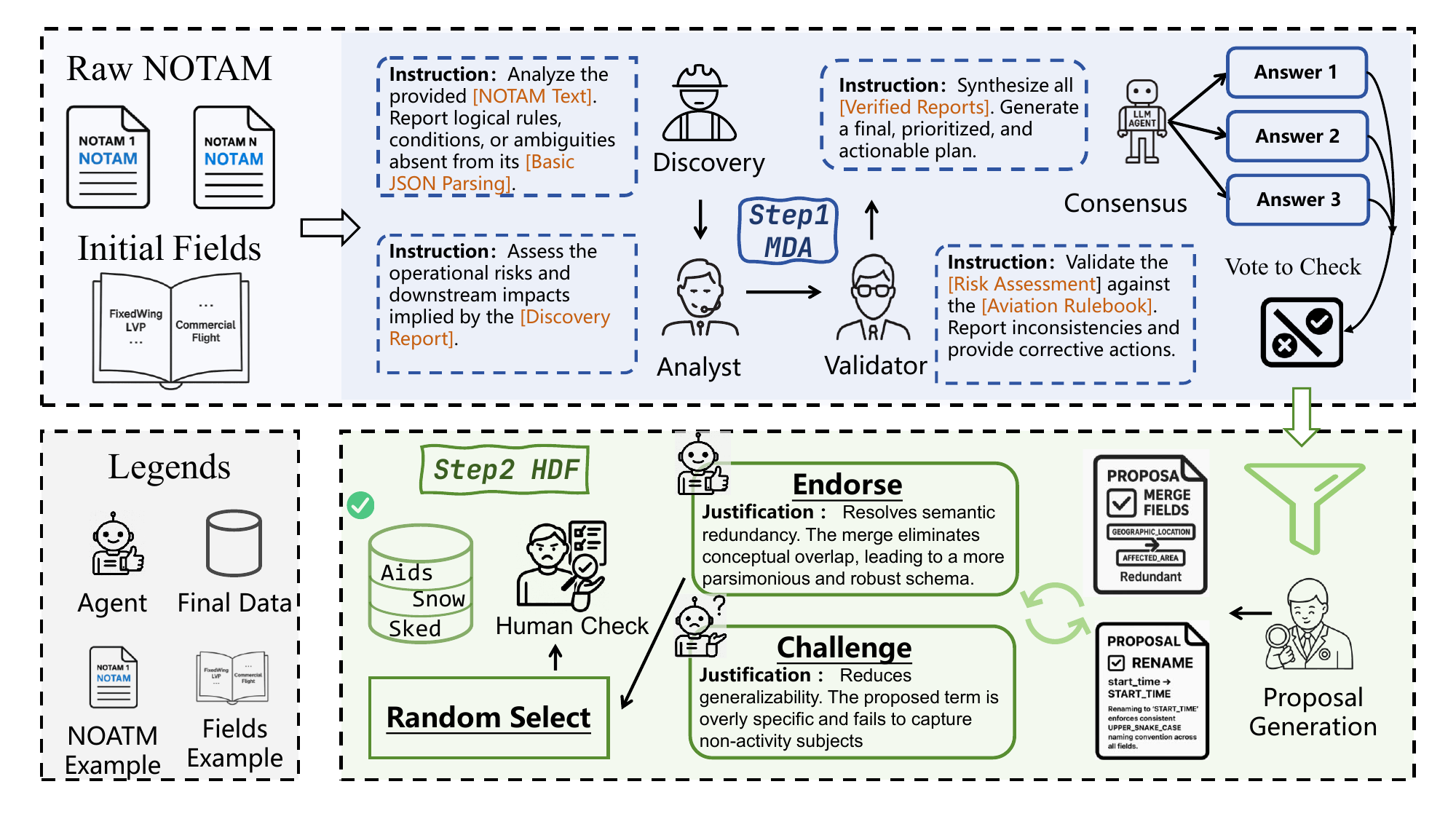}
\caption{Overview of the multi-agent field discovery and refinement framework. The pipeline consists of two main stages: (1) Multi-Agent Field Discovery (MDA) for systematic field extraction, and (2) Hybrid Debate Framework (HDF) for collaborative refinement through structured debate and deterministic consolidation.}
\label{fig:framework}
\end{figure*}

\section{Multi-Agent Framework for Field Discovery and Refinement}
\label{sec:framework}

Precise interpretation of Notices to Air Missions (NOTAMs) is critical to aviation safety. While traditional NOTAMs follow a fixed structure, advanced applications—such as identifying specific restrictions for aircraft models or commercial flights—necessitate the creation of new, finer-grained information fields. This task involves a fundamental trade-off: on one hand, it is essential to discover all potential fields comprehensively to avoid omission of critical information (prioritizing high recall); on the other hand, it demands the rigorous refinement of these fields to ensure their definitions are clear, non-redundant, and hold business value (prioritizing high precision).

The unique grammatical and stylistic characteristics of NOTAMs render traditional phrase-mining methods ineffective. Moreover, employing a single Large Language Model (LLM) often cannot reconcile the competing requirements of exploratory completeness and strict accuracy. To address these challenges, we propose an innovative two-stage multi-agent framework that decomposes the complex task into two key phases, as depicted in Figure~\ref{fig:framework}: Multi-Agent Discovery (MDA) systematically uncovers candidate fields through a structured agent pipeline focused on maximizing recall, while Hybrid Debate Framework (HDF) refines candidate fields via structured debate mechanisms with adversarial critique and deterministic decision rules, aiming to maximize precision and robustness. This ``explore-then-refine'' design establishes a systematic balance between recall and precision, producing a field set of sufficient quality for deployment in practical systems.

\subsection{Problem Formulation and Theoretical Framework}

Given a raw NOTAM text \(t\), the goal is to extract a structured set of fields \(F(t) = \{f_1, f_2, \dots, f_N\}\), where each field \(f_i = (n_i, d_i, v_i, s_i)\) consists of a unique name \(n_i\), a precise description \(d_i\), extracted value \(v_i\), and supporting textual evidence \(s_i\) --- excerpts from the original text \(t\). An ideal framework must achieve high recall (capturing all business-relevant fields without omission), high precision (excluding spurious, redundant, or inconsistently defined fields), and robustness (avoiding premature consensus and errors caused by model biases or groupthink).

Following the \textit{Latent Concept Space} perspective \cite{Estornell2024}, the output of an agent (e.g., a candidate field \(z\)) is generated based on an underlying latent concept \(\theta \in \Theta\) inferred from the context, formally expressed as:
\[
P(z \mid \text{context}, \phi) = \sum_{\theta \in \Theta} P(z \mid \theta, \phi) \, P(\theta \mid \text{context}, \phi)
\]
where \(\phi\) denotes model parameters. However, when multiple agents collaborate using similar mechanisms, two main failure modes emerge: the \textbf{Echo Chamber Effect}, where homogeneous agents rapidly reinforce initial consensus opinions, causing premature convergence and limiting exploration of alternative hypotheses; and the \textbf{Tyranny of the Majority}, where an erroneous concept can dominate if a majority coincidentally endorses it, suppressing correct minority views due to exponentially increasing peer pressure \cite{Estornell2024}. Such effects undermine the quality and robustness of conventional multi-agent methods, necessitating active intervention mechanisms that our Hybrid Debate Framework (HDF) systematically incorporates.

\begin{algorithm}[!ht]
\caption{Multi-Agent Field Discovery and Refinement (MDA-HDF)}
\label{alg:mda-hdf}
\KwIn{NOTAM text $t$, similarity threshold $\tau = 0.7$, max iterations $I_{max} = 5$}
\KwOut{Refined field set $F_{final}$}

\BlankLine
\tcp{Stage 1: Multi-Agent Discovery (MDA)}
$Z_1 \leftarrow$ DiscoveryAgent($t$)\;
$Z_2 \leftarrow$ AnalysisAgent($t, Z_1$)\;
$Z_3 \leftarrow$ ValidationAgent($t, Z_1, Z_2$)\;
$Z_{MDA} \leftarrow$ ConsensusAggregator($Z_3$, $\tau$)\;

\BlankLine
\tcp{Stage 2: Hybrid Debate Framework (HDF)}
$F_{current} \leftarrow Z_{MDA}$\;
$iteration \leftarrow 0$\;

\While{$iteration < I_{max}$ \textbf{and} HasNewProposals()}{
    \BlankLine
    \tcp{Propose: Generate improvement proposals}
    $P_{struct} \leftarrow$ ConsolidationExpert($F_{current}$)\;
    $P_{term} \leftarrow$ TerminologyExpert($F_{current}$)\;
    
    \BlankLine
    \tcp{Critique: Adversarial evaluation}
    $P_{accepted} \leftarrow$ CriticAgent($P_{struct} \cup P_{term}$, $F_{current}$)\;
    
    \BlankLine
    \tcp{Consolidate: Apply accepted proposals}
    $F_{current} \leftarrow$ FieldManager.Apply($P_{accepted}$, $F_{current}$)\;
    
    $iteration \leftarrow iteration + 1$\;
}

\BlankLine
$F_{final} \leftarrow$ PostProcess($F_{current}$)\;
\Return{$F_{final}$}
\end{algorithm}

\subsection{Multi-Agent Discovery (MDA) and Hybrid Debate Framework (HDF)}

The MDA stage maximizes recall by ensuring coverage of all potential fields through a sequential pipeline architecture composed of three agents with distinct roles. Each agent processes and enriches the outputs of the previous, progressively expanding coverage. The \textbf{Discovery Agent} performs a broad, surface-level scan of raw text \(t\) to identify any spans possibly constituting fields, yielding an initial candidate set \(Z_1\). The \textbf{Analysis Agent} enriches \(Z_1\) by leveraging domain knowledge—resolving ambiguities and inferring implicit fields—producing a semantically enhanced set \(Z_2\). The \textbf{Validation Agent} performs consistency checks, deduplicates obvious overlaps, and preliminarily filters for business relevance, yielding a refined candidate set \(Z_3\). 

A \textbf{Consensus Aggregator} then merges semantically similar fields from \(Z_3\), producing the final candidate set \(Z_{\mathrm{MDA}}\) for the refinement stage. The transformation follows: \(Z_{\mathrm{MDA}} = \text{ConsensusAggregator}(Z_3, \tau)\) where \(\tau = 0.7\) is the Jaccard similarity threshold for semantic merging. Quality control mechanisms include semantic similarity thresholds for deduplication, domain relevance scoring based on aviation terminology frequency, and consistency checks across agent outputs using field name normalization.

The HDF stage addresses ambiguities, redundancies, and formulation weaknesses from MDA's output, systematically counteracting echo chamber and tyranny effects through structured intervention. The HDF stage initializes with \(F_{\mathrm{current}}^{(0)} = Z_{\mathrm{MDA}}\) and iteratively refines through the transformation \(F_{\mathrm{current}}^{(t+1)} = \mathrm{HDF}(F_{\mathrm{current}}^{(t)})\). Three core interventions form the foundation: Conceptual Partitioning splits the refinement task into two orthogonal subtasks (structural optimization for merging/deduplication and expressive optimization for naming/description), each handled by separate expert agents to foster diversity and prevent dominance by any single concept. Adversarial Critique employs an independent Critic Agent to challenge proposals with counterarguments, reducing acceptance of flawed concepts through modification intervention that influences agents' posterior beliefs. Deterministic Consolidation executes the final application of proposals via the programmatic FieldManager using deterministic rules rather than probabilistic voting, ensuring transparent, reproducible decisions while avoiding tyranny effects from majority overrides.

The refinement process follows a systematic three-stage cycle. During the Propose stage, two expert agents generate improvement proposals in parallel: the Consolidation Expert proposes merges or removals to optimize structure, while the Terminology Expert proposes renaming or description clarifications. The Critique stage involves the independent Critic Agent reviewing all proposals, labeling each as supported or challenged with detailed justification. Finally, the Consolidate stage sees the FieldManager deterministically accepting all unchallenged proposals, ensuring efficient and stable progress. The HDF cycle terminates when no new proposals are generated for two consecutive iterations, or when a maximum of 5 iterations is reached to prevent infinite loops. Conflict resolution applies a priority system where structural consolidation proposals take precedence over terminological changes, and proposals with stronger textual evidence are preferred. Final field sets undergo post-processing validation including duplicate detection, completeness checking against aviation standards, and format normalization for downstream applications.

This two-stage "explore-then-refine" architecture is intentionally designed to ensure both robustness and interpretability. The MDA stage prioritizes high recall, which may introduce some noise or initial misinterpretations. It is worth noting that unlike in open-domain question answering, "hallucinations" in our structured extraction task manifest as outputs that are unfaithful to the source text rather than ungrounded factual fabrications \cite{Huang2023ASO}. The subsequent HDF stage, particularly the Critic Agent, is specifically designed to mitigate such issues by systematically filtering these imperfections through structured debate and adversarial critique. This design prevents the propagation of initial errors, a claim substantiated by our ablation study (Table~\ref{tab:field_discovery_results}), where removing HDF caused a significant drop in precision from 92.0\% to 84.0\%. Furthermore, while the framework is complex, it remains debuggable. Each agent operates modularly with structured, recordable outputs (e.g., JSON proposals). This allows for a clear audit trail, making it straightforward to trace errors back to the specific agent and stage where they originated.

\subsection*{Lemma 1 (Conceptual Space Expansion in MDA)}

Let \(\Theta(Z)\) denote the latent concept set covered by a candidate field collection \(Z\), and define \(\mu(\cdot)\) as a monotonic set function measuring conceptual coverage, satisfying \(\mu(A \cup B) \geq \max(\mu(A), \mu(B))\) for any concept sets \(A, B\). Under the assumption that agents possess complementary knowledge (\(\phi_{\mathrm{discover}} \neq \phi_{\mathrm{analyst}} \neq \phi_{\mathrm{validator}}\)), the final candidate set \(Z_{\mathrm{MDA}}\) produced via the pipeline achieves equal or superior conceptual coverage compared to any individual agent output:
\[
\mu(\Theta(Z_{\mathrm{MDA}})) \geq \max_{k \in \{1,2,3\}} \mu(\Theta(Z_k))
\]
where the measure \(\mu(\cdot)\) quantifies the conceptual "breadth" or "diversity" of a set, with monotonicity ensuring subsets cannot exceed supersets in coverage.

\subsection*{Theorem 1 (Robustness of the Hybrid Debate Framework)}

Let \(\Omega\) be the sample space of all possible NOTAM texts and field extraction scenarios. Denote by \(\mathcal{P}_{\mathrm{HDF}}(\theta^* \mid t)\) the probability that HDF converges to the correct latent concept \(\theta^*\) given input \(t\), and by \(\mathcal{P}_{\mathrm{Vanilla}}(\theta^* \mid t)\) the analogous probability for a conventional voting-based debate. Adversarial critique decreases the likelihood of accepting faulty proposals: \(P_{\mathrm{reject}}(z_{\mathrm{proposal}} \mid z_{\mathrm{critique}}) > P_{\mathrm{reject}}(z_{\mathrm{proposal}})\), while partitioning and deterministic rules significantly reduce convergence to incorrect majority concepts \(\theta'_{\mathrm{maj}}\): \(\mathcal{P}_{\mathrm{HDF}}(\text{converge to } \theta'_{\mathrm{maj}} \mid t) < \mathcal{P}_{\mathrm{Vanilla}}(\text{converge to } \theta'_{\mathrm{maj}} \mid t)\).

The ConsensusAggregator implements semantic merging via Jaccard similarity: 
\[
\text{merge}(f_i, f_j) \iff \text{Jaccard}(f_i, f_j) > \tau
\]
where $\tau = 0.7$ is the similarity threshold. The FieldManager applies deterministic rules:
\begin{align*}
\text{Apply}(P, F) &= \{f \in F : \neg \exists p \in P, \text{conflicts}(p, f)\} \\
&\quad \cup \{p \in P : \text{unchallenged}(p)\}
\end{align*}

\section{NOTAM Parsing Methodology}
\label{sec:methodology}

This section presents our comprehensive approach to NOTAM parsing, focusing on systematic prompt engineering strategies and domain-specific optimization techniques. We detail the evaluation of foundational baselines through advanced reasoning techniques, followed by an exploratory post-processing method designed to address specific, persistent extraction errors. Our methodology is grounded in a core principle: parsing in the aviation domain requires careful consideration of specialized terminology, structured output formats, and the inherent complexity of safety-critical information.

\subsection{Core Prompting Strategies}
\label{sec:core_prompting}

We systematically explored multiple prompting paradigms to optimize Large Language Model (LLM) performance on NOTAM parsing tasks, with particular attention to the trade-off between reasoning complexity and extraction accuracy in aviation contexts.

\paragraph{Zero-shot and Few-shot In-Context Learning (ICL)}
\label{sec:icl}

Our baseline approach employs carefully crafted zero-shot prompts that leverage pre-trained knowledge while incorporating aviation domain-specific context. The zero-shot strategy focuses on clear instruction design, including explicit field definitions and output format specifications that align with ICAO standards and operational requirements.

For few-shot In-Context Learning (ICL), we adopt a 5-shot approach with strategically selected examples that demonstrate domain-specific pattern learning for professional NOTAM terminology and abbreviation handling (e.g., RWY 09L CLSD $\rightarrow$ interpreted as runway closure), provide clear JSON output structure to reduce parsing errors and ensure field extraction consistency, and cover complex mixed-content NOTAMs and scenarios requiring heavy inference. These ICL examples are vetted through review by professional aviation dispatchers to ensure representative coverage across different Q-code categories and operational scenarios. Each example includes the original NOTAM text and its corresponding structured output, emphasizing semantic inference rather than literal text matching. Detailed examples are available in our code repository.

\paragraph{Advanced Reasoning and Reliability Strategies}
\label{sec:advanced_reasoning}

To address more complex NOTAMs requiring multi-step inference and domain expertise, we investigated advanced reasoning techniques tailored specifically for aviation contexts.

Our Chain-of-Thought (CoT) reasoning implementation employs step-by-step reasoning prompts designed for scenarios requiring complex spatial-temporal decomposition and implicit information inference. This approach focuses on decomposing regional restrictions with mixed content that require systematic analysis of spatial and temporal relationships, and inferring implicit information from aviation context where domain knowledge is essential for correct interpretation (e.g., ``RWY 09L degraded 300M'' implies a BALS downgrade based on approach lighting system specifications). The goal was to test whether explicit reasoning steps could improve accuracy on these challenging cases that require domain expertise beyond simple text extraction.

While information extraction typically employs deterministic temperature settings of $0.0$ for consistent outputs, we explored self-consistency as a potential reliability enhancement mechanism. This investigation involved generating multiple outputs using varied temperature settings ($0.3$, $0.7$, $1.0$) and employing majority voting to determine the final extraction results. This approach tested the hypothesis that consensus across multiple reasoning paths could yield more robust results, particularly for ambiguous cases where multiple valid interpretations might exist.

\subsection{Exploratory Approach: Selective Refinement via Contrastive Validation (SRCV)}
\label{sec:srcv_method}

Recognizing that even the most sophisticated prompting strategies can fail on specific, complex cases involving nuanced aviation terminology or implicit reasoning requirements, we conducted a preliminary investigation into a post-processing method we term \emph{Selective Refinement via Contrastive Validation (SRCV)}. SRCV is designed as a two-step error-correction loop, intended to be applied selectively to fields with low confidence scores or known systematic errors after an initial ICL-based extraction.

The SRCV process begins with an identification phase where specific fields in the JSON output are identified as candidates for refinement based on confidence scoring or systematic error patterns observed during development. The targeted validation phase then employs a second, specialized prompt sent to the LLM, presenting the original NOTAM text alongside the extracted value for the specific field under review. The model is asked to validate or correct the extraction, with the prompt augmented by contrastive examples or explicit domain rules to guide the refinement process (e.g., ``Validate the end time. Remember that if `PERM' is present, the end time should be null.''). This approach is considered exploratory, and our primary aim was to assess its potential for correcting specific, stubborn errors that resist standard prompting approaches, particularly those involving complex temporal relationships or implicit domain knowledge requirements.

\subsection{Methodological Validation Framework}
\label{sec:validation_framework}

Our methodology development follows a systematic validation approach that ensures robustness and practical applicability across diverse NOTAM processing scenarios. We conduct comprehensive comparisons between zero-shot, 5-shot ICL, CoT, and self-consistency approaches using consistent evaluation metrics and controlled experimental conditions, as presented in Table~\ref{tab:results_textwidth}. 

Our evaluation spans multiple model architectures, including Qwen3-8B and GPT-4.1-Nano, Gemini-2.5-Flash, DeepSeek V3.2, and GPT-5-Nano, to ensure methodology robustness across different LLM capabilities and scales. This multi-model approach reveals important insights about the relationship between model size and prompt sensitivity, particularly in aviation domain applications where specialized terminology and implicit reasoning requirements may affect smaller and larger models differently. The systematic evaluation framework provides concrete guidance for practitioners seeking to implement robust NOTAM parsing systems in operational environments, establishing a foundation for reliable automated processing of aviation safety communications.

This comprehensive methodology provides a systematic framework for applying Large Language Models to aviation safety-critical information extraction while maintaining the precision and reliability requirements essential for operational aviation systems. The combination of systematic prompt engineering strategies with domain-specific optimizations establishes a robust foundation for automated processing of aviation safety communications.

\sisetup{parse-numbers=false}
\newcommand{\redbf}[1]{\textcolor{red}{\textbf{#1}}} % 红色+加粗（模型名用）
\newcommand{\red}[1]{\textcolor{red}{#1}}            % 红色但不加粗（数值用）

\begin{table*}[!ht]
\centering
\small
\caption{Performance comparison of various methods on five datasets. Values indicate F1 scores (\%). The best-performing result is highlighted in bold.}
\label{tab:results_textwidth}
\setlength{\tabcolsep}{4pt}
\renewcommand{\arraystretch}{1.05}
\begin{tabularx}{0.98\textwidth}{
@{} >{\centering\arraybackslash}p{2.2cm}
>{\RaggedRight\arraybackslash}p{1.8cm}
>{\RaggedRight\arraybackslash}X
*{5}{>{\centering\arraybackslash}p{1.7cm}} @{}}
\toprule
\textbf{Setup} & \textbf{Method} & \textbf{Backbone}
& {\makecell[l]{Airspace\\Management}}
& {\makecell[l]{Ground\\Facility}}
& {\makecell[l]{Landing\\Aid}}
& {\makecell[l]{Runway \&\\Taxi-way}}
& {\makecell[l]{Flight\\Hazard}} \\
\cmidrule(r){4-8}
& &
& {(4286)} & {(2928)} & {(1902)} & {(1335)} & {(1896)} \\
\midrule
\multirow[c]{2}{*}{\textbf{Traditional}}
& Regex-based & {--} & 36.5 & 37.2 & 35.8 & 36.1 & 34.9 \\
& UIE & {--} & 51.3 & 52.1 & 50.7 & 51.5 & 49.8 \\
\midrule
\multirow[c]{15}{*}{\textbf{LLM-based}}
& \multirow[c]{5}{*}{Zero-shot} 
& GPT4.1-Nano & 58.5 & 91.0 & 95.5 & 75.8 & 55.0 \\
& & Qwen3-8B & 67.8 & 90.8 & 95.6 & 82.4 & 64.3 \\
& & Gemini-2.5-Flash & 68.5 & 90.9 & 95.8 & 80.5 & 66.2 \\
& & DeepSeek V3.2 & 67.0 & 90.7 & 95.7 & 79.8 & 64.3 \\
& & GPT-5-Nano & 66.2 & 90.8 & 95.6 & 79.2 & 63.5 \\
\cmidrule(lr){2-8}
& \multirow[c]{5}{*}{CoT}
& GPT4.1-Nano & 63.7 & 90.4 & 95.6 & 72.4 & 73.3 \\
& & Qwen3-8B & 67.1 & 90.7 & 95.9 & 78.6 & 62.5 \\
& & Gemini-2.5-Flash & 72.5 & 91.1 & 96.0 & 82.1 & 72.8 \\
& & DeepSeek V3.2 & 71.2 & 90.9 & 95.8 & 81.7 & 71.1 \\
& & GPT-5-Nano & 70.5 & 91.0 & 95.7 & 81.0 & 70.4 \\
\cmidrule(lr){2-8}
& \multirow[c]{5}{*}{ICL (5-shot)}
& GPT4.1-Nano & 69.2 & 91.1 & 96.0 & 79.9 & 81.7 \\
& & Qwen3-8B & 71.1 & 91.6 & 95.9 & 81.0 & 83.8 \\
& & \textbf{Gemini-2.5-Flash} & \textbf{77.4} & \textbf{91.7} & \textbf{96.3} & \textbf{84.2} & \textbf{84.3} \\
& & DeepSeek V3.2 & 76.0 & 91.5 & 96.1 & 83.6 & 82.3 \\
& & GPT-5-Nano & 75.3 & 91.3 & 96.0 & 82.9 & 81.5 \\
\bottomrule
\end{tabularx}
\end{table*}

\section{Experiments}
\label{sec:experiments}

This section presents a systematic evaluation of our proposed NOTAM semantic parsing framework through two core experiments. The first experiment evaluates the capability of our multi-agent collaboration framework (MDA-HDF) to automatically identify novel and critical information fields without predefined field schemas, thereby extending the scope of traditional NOTAM information extraction tasks. The second experiment focuses on the NOTAM parsing task, conducting rigorous comparative analysis of various Large Language Model (LLM)-based prompting strategies.

\subsection{Field Discovery Evaluation}

\begin{table*}[!ht]
\centering
\small
\caption{Comparison of field extraction results from different methods on the same input NOTAM text}
\label{tab:example_output_comparison}
\begin{tabularx}{\textwidth}{
  *{6}{>{\raggedright\arraybackslash}X}  % 六列全部自适应
}
\toprule
\textbf{Input NOTAM} & \textbf{spaCy} & \textbf{RAKE} & \textbf{TF-IDF} & \textbf{AutoPhrase} & \textbf{MDA with HDF} \\
\midrule
E) RWY 12L/30R CLSD DUE TO MAINT AND STRONG WIND. &
12L/30 clsd.due maint.and strong wind &
12L/30 cls.strong wind &
12L/30R clsd maint &
Runway Closure Maintenance &
Runway Closure Maintenance WIND \\
\bottomrule
\end{tabularx}
\end{table*}

We validate the capability of our MDA-HDF framework to automatically discover novel, low-frequency yet operationally critical information fields from raw NOTAM text, formulating this as a phrase mining problem. It is crucial to note that this framework is designed as an offline tool for dataset enhancement—a one-time process to generate high-quality field definitions. It is not intended for real-time operational deployment, and thus its evaluation focuses on output quality (precision and recall) rather than inference latency. Due to the unique linguistic structures and domain-specific expressions in NOTAM text, traditional phrase or keyword extraction methods struggle to adapt to their diverse and information-sparse characteristics, resulting in significantly limited performance. We used 500 randomly sampled NOTAM instances from the Knots dataset with expert-annotated ground truth labels, comparing against representative baseline methods including spaCy noun phrase extraction, RAKE keyword extraction, TF-IDF statistical methods, AutoPhrase phrase mining, and single LLM approaches. Our proposed methods include the complete MDA-HDF framework and an ablation variant (MDA w/o HDF) that removes the Hybrid Debate Framework to assess the specific contribution of the HDF module. Performance evaluation uses Precision, Recall, and F1-score metrics. All LLM-based methods employ the same model version (gpt4.1-nano), with result stability controlled through multiple runs. To clarify the implementation, each agent within our framework is simulated by making a distinct API request with a role-specific prompt to a single LLM endpoint. This multi-step process is an offline, one-time procedure for dataset enhancement and does not represent a real-time deployment architecture.

As shown in Table~\ref{tab:field_discovery_results}, the MDA-HDF framework achieves an F1-score of 92\%, significantly outperforming traditional phrase mining methods and further improving upon single LLM approaches. Traditional methods, relying on word frequency statistics or general grammatical rules, fail to effectively adapt to the highly specialized and low-frequency critical information expressions in NOTAMs, resulting in severely imbalanced precision or recall. For example, spaCy achieves high recall (0.92) but extremely low precision (0.18), demonstrating poor noise filtering capability; AutoPhrase shows higher precision (0.85) but severely insufficient recall (0.34), resulting in incomplete field extraction. In contrast, single LLM methods demonstrate stable performance with F1 scores typically fluctuating between 0.85 and 0.96, reflecting the significant advantages of language models in understanding complex semantics and context, though they still suffer from insufficiently refined candidate field filtering and occasional redundancy.

The ablation study reveals that MDA without the HDF module achieves recall of 96\%, demonstrating the capability of multi-agent collaboration strategies to discover comprehensive information; however, precision drops to 84\%, indicating that unconstrained candidate generation tends to produce more noisy fields. After introducing the HDF module, multi-round hybrid debate and semantic integration significantly improve field extraction accuracy to 92\% while maintaining high recall, ultimately achieving a good balance between precision and recall and substantially improving overall F1 performance.

For intuitive comparison, Table~\ref{tab:example_output_comparison} presents extracted key fields from the same typical NOTAM text using different methods. Traditional tools produce verbose output containing numerous irrelevant terms, while single LLM generates semantically complete and relatively concise fields but with some missing or ambiguously expressed key information. The complete MDA-HDF output not only covers important low-frequency information but also ensures accurate field descriptions and diversity, demonstrating the framework's advantages in semantic understanding and information integration.

In summary, traditional phrase extraction methods, due to their over-reliance on statistical features and general language patterns, cannot effectively capture the rich and specialized low-frequency key fields in NOTAM text. Single LLM methods already demonstrate strong semantic understanding capabilities, achieving relatively high precision and recall. The MDA-HDF framework, based on multi-agent collaboration and hybrid debate mechanisms, further improves information extraction accuracy and completeness through structured adversarial semantic integration, better meeting the requirements for automated critical information discovery in the NOTAM domain.

\begin{table}[!ht]
\small
\centering
\caption{Field discovery performance. F1 denotes overall extraction quality. 
$\uparrow$ means higher is better, $\downarrow$ means lower is better.}
\label{tab:field_discovery_results}
\begin{tabularx}{0.46\textwidth}{l *{3}{S[table-format=2.2]} S[table-format=2.1]}
\toprule
\textbf{Method} & \textbf{Prec. $\uparrow$} & \textbf{Rec. $\uparrow$} & \textbf{F1 $\uparrow$} & \textbf{Avg $\downarrow$} \\
\midrule
\multicolumn{5}{l}{\textit{Traditional baselines}} \\
spaCy (Noun Phr.) & 18.0 & \textbf{92.0} & 30.0 & 47.2 \\
RAKE (Keyword) & 45.0 & 78.0 & 57.0 & 18.6 \\
TF-IDF & 41.0 & 69.0 & 51.0 & 15.3 \\
AutoPhrase & 85.0 & 34.0 & 48.0 & \textbf{3.7} \\
\midrule
\multicolumn{5}{l}{\textit{Single LLM}} \\
Single LLM & 89.0 & 83.0 & 86.0 & 9.7 \\
\midrule
\multicolumn{5}{l}{\textit{Proposed methods}} \\
MDA (w/o HDF) & 84.0 & \textbf{96.0} & 89.0 & 14.2 \\
MDA-HDF (Full) & \textbf{92.0} & 93.0 & \textbf{92.0} & 9.3 \\
\bottomrule
\end{tabularx}
\end{table}

\subsection{NOTAM Parsing Evaluation}

We conduct systematic evaluation of the NOTAM parsing task using the Knots dataset with standard 80/10/10 splits, reporting strict entity-level metrics requiring exact matches for both field types and normalized values. Our evaluation encompasses comprehensive analysis of various prompting strategies, domain-specific optimizations, and exploratory refinement techniques across multiple model architectures.

\paragraph{Experimental Configuration and Baseline Comparison}
As shown in Table~\ref{tab:results_textwidth}, we compare the performance of various methods on the Knots dataset across five major operational domains. Considering practical application requirements, we adopt five representative models: Qwen3-8B, GPT4.1-Nano, Gemini-2.5-Flash, DeepSeek V3.2, and GPT-5-Nano as backbones, ensuring robust evaluation across different model scales and capabilities.

Traditional methods demonstrate limited effectiveness in NOTAM parsing tasks. Regex-based heuristic approaches maintain F1-scores around 36-37\% across all domains, reflecting the challenge of capturing semantic nuances through rule-based pattern matching. The UIE method performs moderately better with F1-scores in the 50-52\% range, yet still falls short of the semantic understanding required for complex aviation terminology processing.

LLM-based methods reveal clear performance hierarchies and domain-specific variations. Zero-shot prompting strategies achieve baseline F1-scores ranging from 55-96\%, with notable performance disparities across operational domains. Ground Facility and Landing Aid domains show consistently high performance ($>90$)\% across all models, likely due to their more standardized terminology and clearer structural patterns. In contrast, Airspace Management and Flight Hazard domains present greater challenges, with F1-scores varying significantly between 55-84\%, reflecting the complexity of spatial-temporal reasoning and hazard interpretation requirements.

\paragraph{In-Context Learning Superiority and Strategy Effectiveness}
The 5-shot In-Context Learning (ICL) strategy demonstrates consistent superiority across all experimental conditions. As shown in Table~\ref{tab:results_textwidth}, ICL achieves the highest F1-scores across all domains, with Gemini-2.5-Flash reaching up to 96.3\% and consistently outperforming other models. This represents substantial improvements over zero-shot approaches. This superiority can be attributed to several key factors that align particularly well with aviation domain requirements.

Domain-specific pattern learning emerges as a critical advantage of ICL. Aviation NOTAM processing benefits significantly from concrete examples that demonstrate specialized terminology handling, abbreviation normalization, and implicit semantic inference. For instance, examples showing how "RWY 09L CLSD DUE TO MAINT" should be interpreted as a runway closure with maintenance causation provide crucial contextual guidance that enables models to handle similar patterns effectively.

Format consistency represents another substantial benefit of ICL implementation. The inclusion of structured JSON output examples significantly reduces parsing errors and ensures field extraction consistency across different NOTAM formats. This proves particularly valuable in aviation contexts where output reliability and standardization are paramount for operational safety.

Edge case coverage through strategic example selection enables robust handling of complex mixed-content NOTAMs that require heavy inference. Our carefully curated 5-shot examples span different Q-code categories and operational scenarios, providing comprehensive coverage of challenging cases that might otherwise result in extraction failures.

\paragraph{Chain-of-Thought Analysis and Limitations}
Contrary to expectations based on general NLP literature, Chain-of-Thought (CoT) reasoning shows mixed effectiveness in aviation domain parsing. While CoT achieves F1-scores of 62-96\% across all models, with Gemini-2.5-Flash showing strong performance, performance improvements over zero-shot baselines are inconsistent across domains and models.

The mixed results reveal important insights about reasoning strategy application in structured extraction tasks. CoT proves beneficial for complex reasoning scenarios requiring multi-step spatial-temporal decomposition, such as analyzing regional airspace restrictions with overlapping temporal boundaries. However, for more straightforward extraction tasks involving standardized aviation terminology, the additional reasoning steps often introduce unnecessary complexity without corresponding accuracy gains.

Model dependency emerges as a significant factor affecting CoT effectiveness. The reasoning strategy shows more pronounced improvements on GPT4.1-Nano and Gemini-2.5-Flash for certain domains (notably Flight Hazard, improving from 55\% to 73\% and 66.2\% to 72.8\% respectively), while other models demonstrate more modest and inconsistent gains. This suggests that CoT benefits may be more closely tied to underlying model reasoning capabilities than previously anticipated.

\begin{figure}[!ht]
    \centering
    \includegraphics[width=0.49\textwidth]{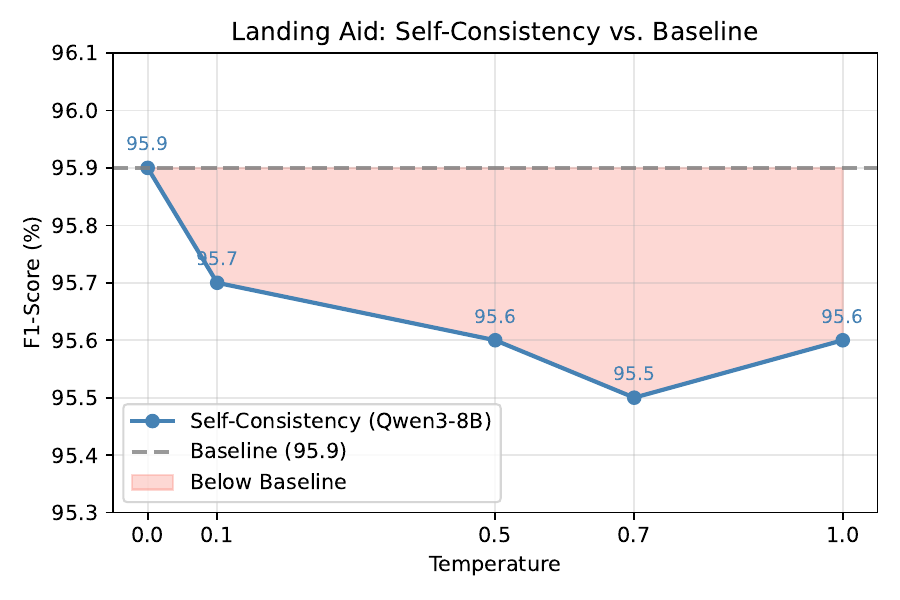}
    \caption{Self-consistency F1-scores vs. temperature on Landing Aid task using Qwen3-8B model.}
    \label{fig:sc}
\end{figure}

\paragraph{Temperature Analysis and Self-Consistency Effects}
Our systematic evaluation of temperature effects on extraction performance reveals critical insights about optimal generation strategies for safety-critical domains. As illustrated in Figure~\ref{fig:sc}, deterministic settings (temperature=0.0) consistently outperform stochastic sampling approaches across all models and domains, with F1-scores showing steady degradation as temperature increases.

The self-consistency analysis using the Qwen3-8B model on the Landing Aid task demonstrates that performance peaks at temperature=0.0 with an F1-score of approximately 95.9\%, then gradually decreases as temperature increases to 0.3, 0.7, and 1.0. This pattern confirms that the precision requirements of safety-critical information processing clearly favor deterministic generation over diversity-oriented approaches.

This finding has critical implications for aviation applications where consistency and reproducibility are paramount. These results confirm that operational NOTAM parsing systems should prioritize reliability over creative interpretation.

\begin{figure}[!t] % 用 figure 而不是 figure*
    \centering
    \includegraphics[width=0.49\textwidth]{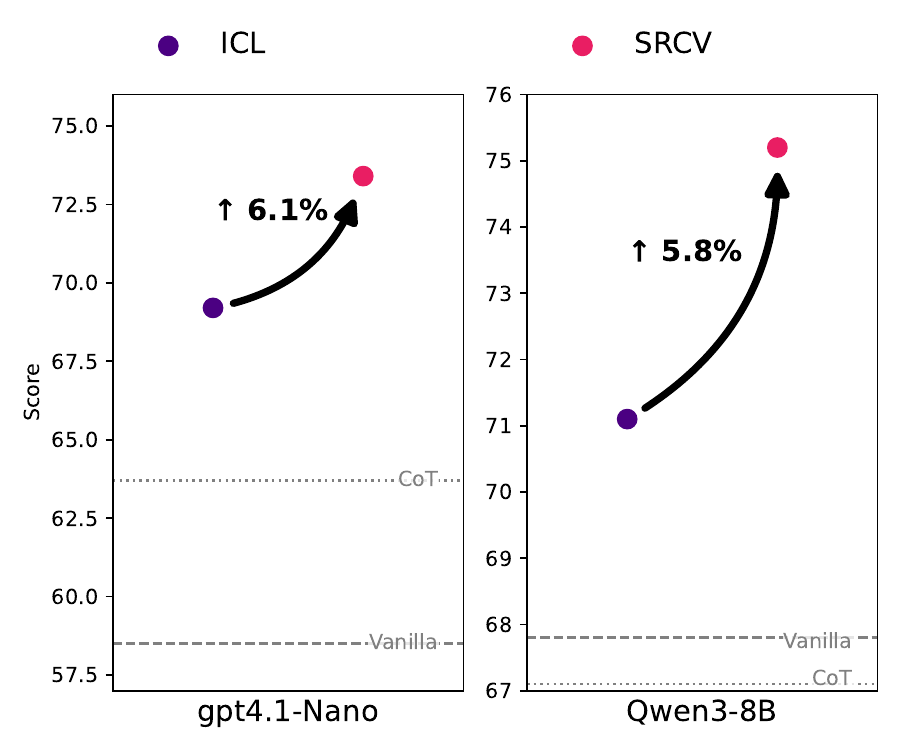}
    \caption{Performance improvement from ICL to SRCV method across different models, showing percentage gains and baseline comparisons.}
    \label{fig:srcv}
\end{figure}

\paragraph{Selective Refinement via Contrastive Validation (SRCV) Analysis}
Our preliminary investigation of Selective Refinement via Contrastive Validation (SRCV) reveals both promising potential and current limitations. As illustrated in Figure~\ref{fig:srcv}, SRCV demonstrates targeted improvements in specific challenging scenarios while introducing risks in others.

The SRCV method shows notable performance improvements when applied to both GPT4.1-Nano and Qwen3-8B models, with improvements varying across different operational domains. For GPT4.1-Nano, SRCV achieves improvements ranging from approximately 2-8\% across different categories, while Qwen3-8B shows similar patterns with improvements of 3-10\% in certain domains. The most significant improvements are observed in complex reasoning scenarios involving temporal relationships and implicit domain knowledge requirements.

However, the exploratory nature of SRCV becomes apparent through mixed results across different field types. While improvements are observed in specific error-prone fields, the approach occasionally introduces unintended side effects in previously accurate extractions. This suggests that SRCV requires more sophisticated targeting mechanisms and careful validation protocols before deployment in operational systems. The preliminary results indicate that selective refinement techniques represent a promising research direction but require substantial development to ensure both safety and reliability in aviation applications.

Targeted effectiveness emerges as SRCV's primary strength, with notable improvements observed in complex temporal relationship parsing and implicit domain knowledge requirements. For cases involving ambiguous time specifications or technical terminology requiring aviation expertise, the contrastive validation approach successfully corrects systematic errors that resist standard prompting strategies. Future work should focus on developing more precise targeting criteria and robust validation mechanisms to minimize collateral effects while maximizing correction effectiveness.

\begin{table}[htbp]
\centering
\caption{Model performance with grouped headers.}
\label{tab:field_semantics_impact}
\begin{tabularx}{0.45\textwidth}{l 
                S[table-format=2.1] S[table-format=2.1] 
                S[table-format=2.1] S[table-format=2.1]}
\toprule
\multirow{2}{*}{Model} & \multicolumn{2}{c}{vanilla} & \multicolumn{2}{c}{icl} \\
\cmidrule(r){2-3} \cmidrule(l){4-5}
                       & {weak} & {clear} & {weak} & {clear} \\
\midrule
Qwen3-8B    & 95.5 & 95.5 & 95.6 & 95.9 \\
Qwen3-14B   & 95.9 & 96.1 & 95.8 & 96.2 \\
Qwen3-32B   & 96.0 & 96.1 & 96.1 & 96.3 \\
GPT4.1-Nano & 95.5 & 95.9 & 96.0 & 96.2 \\
GPT4.1-Mini & 95.9 & 96.2 & 96.0 & 96.4 \\
GPT4.1-Full & 94.8 & 94.8 & 95.1 & 95.6 \\
\bottomrule
\end{tabularx}
\end{table}

\paragraph{Domain-Specific Optimization Impact}
Our investigation into field naming conventions reveals substantial impact on extraction performance, particularly for smaller models. Experiments comparing generic field names (e.g., "upper", "lower") with descriptive, domain-aligned alternatives (e.g., "height\_upper\_limit", "height\_lower\_limit") demonstrate consistent F1-score improvements of 0.2--0.4\% for models like GPT-4.1-Nano, as detailed in Table~\ref{tab:field_semantics_impact}.

The performance improvements from semantic clarity are most pronounced in smaller models, confirming our hypothesis about the relationship between model scale and prompt sensitivity. GPT4.1-Nano shows +0.4\% improvement in vanilla settings and +0.2\% in ICL settings, while larger models demonstrate diminishing sensitivity to field naming conventions. This finding has important implications for practical deployment scenarios where computational constraints may necessitate smaller model usage.

Domain alignment with standard aviation terminology proves consistently beneficial across all model scales. Field names that correspond directly to ICAO specifications and operational terminology (e.g., "TORA" instead of "declared\_distance") leverage models' pre-trained aviation knowledge more effectively, resulting in improved semantic understanding and reduced ambiguity in extraction tasks.

\begin{table}[!ht]
\centering
\small
\caption{Complexity analysis of parsing tasks across NOTAM categories. The instruction count reflects the number of explicit rules, conditional logic statements, and mapping constraints defined in each prompt.}
\label{tab:complexity_analysis}  % ← 整个表格都变红
\begin{tabularx}{0.45\textwidth}{
  l
  S[table-format=3.1]
  S[table-format=2.0]
}
\toprule
\textbf{Category} & {\textbf{Avg. Length}} & {\textbf{Avg. Instructions}} \\
\midrule
Airspace Mgmt.  & 294.9 & 7.2 \\
Ground Facility & 67.3  & 4.3  \\
Landing Aid     & 84.6  & 4.7 \\
Runway/Taxiway  & 353.7 & 6.4 \\
Flight Hazard   & 318.2 & 6.6 \\
\bottomrule
\end{tabularx}
\end{table}

To provide a deeper analysis of the performance disparities observed in Table~\ref{tab:results_textwidth}---particularly the lower F1 scores in the ``Airspace Management'' and ``Flight Hazard'' domains---we quantified the complexity of the parsing tasks for each category, following the methodology of prior work~\cite{zhang2025iopo}. As detailed in Table~\ref{tab:complexity_analysis}, this analysis was based on two key metrics: the average length of the input NOTAMs and the average number of instructions the model must follow (i.e., rule constraints, format constraints, and value constraints). The latter is derived from our prompt design, and an average is used because the number of applicable constraints varies even for NOTAMs within the same category depending on the information they contain. The data clearly indicates that the ``Airspace Management'' and ``Flight Hazard'' tasks are significantly more complex, involving not only longer inputs but also a greater number of instructions to follow (7.2 and 6.6, respectively). This heightened complexity directly correlates with the observed performance gap, providing a clear, data-driven explanation for the challenges these domains present.

\paragraph{Parameter Analysis}
To ensure the robustness and optimal performance of our proposed methods, we conducted targeted parameter analyses for key hyperparameters. Table~\ref{tab:jaccard_sensitivity} presents a sensitivity analysis for the Jaccard similarity threshold used in the Consensus Aggregator of our MDA framework. The results demonstrate a clear trade-off between precision and recall. A lower threshold (e.g., 0.5) increases recall by merging more fields but harms precision by incorrectly combining dissimilar ones. Conversely, a higher threshold (e.g., 0.9) improves precision at the cost of recall. The F1-score, which balances both metrics, peaks at a threshold of 0.7, validating our choice for this parameter. Similarly, to justify the selection of 5-shot for our ICL strategy, we performed an ablation study, with the results shown in Table~\ref{tab:icl_shot_ablation}. The performance across all backbone models generally improves as the number of shots increases from one to five. However, the gains diminish and performance saturates at 5-shot, with 7-shot offering no significant additional benefit. Therefore, 5-shot was chosen as the optimal configuration, providing the best balance between accuracy and computational efficiency.

\begin{table}[!ht]
\centering
\small
\caption{Sensitivity analysis of the similarity threshold for the field merging task. 
Precision and F1-score achieve the best trade-off around a threshold of 0.7, where the F1-score reaches its maximum (92.5\%).}
\label{tab:jaccard_sensitivity}
\begin{tabularx}{0.45\textwidth}{c
                                 >{\centering\arraybackslash}X
                                 >{\centering\arraybackslash}X
                                 >{\centering\arraybackslash}X}
    \toprule
    Threshold & Prec. (\%) & Rec. (\%) & F1 (\%) \\
    \midrule
    0.5 & 85.2 & \textbf{96.1} & 90.3 \\
    0.6 & 89.8 & 94.5 & 92.1 \\
    0.7 & 92.0 & 93.0 & \textbf{92.5} \\
    0.8 & 95.3 & 88.2 & 91.6 \\
    0.9 & \textbf{97.1} & 81.5 & 88.6 \\
    \bottomrule
\end{tabularx}
\end{table}

\begin{table}[!ht]
\centering
\small
\caption{Ablation study on the number of examples for In-Context Learning (ICL) on the Runway \& Taxiway task. The table shows F1-scores for different numbers of shots. Performance generally saturates at 5-shot, which offers the best trade-off between accuracy and efficiency.}
\label{tab:icl_shot_ablation}
\begin{tabularx}{0.45\textwidth}{
  l
  >{\centering\arraybackslash}X
  >{\centering\arraybackslash}X
  >{\centering\arraybackslash}X
  >{\centering\arraybackslash}X
}
\toprule
\multirow{2}{*}{\textbf{Backbone}} & \multicolumn{4}{c}{\textbf{F1-Score (\%)}} \\
\cmidrule(lr){2-5}
& {\textbf{1-shot}} & {\textbf{3-shot}} & {\textbf{5-shot}} & {\textbf{7-shot}} \\
\midrule
GPT4.1-Nano      & 75.2 & 78.8 & 79.9 & 79.8 \\
Qwen3-8B         & 76.5 & 80.1 & 81.0 & 81.1 \\
Gemini-2.5-Flash & 79.1 & 83.0 & \textbf{84.2} & \textbf{84.2} \\
DeepSeek V3.2    & 78.5 & 82.4 & 83.6 & 83.5 \\
GPT-5-Nano       & 77.8 & 81.7 & 82.9 & 82.8 \\
\bottomrule
\end{tabularx}
\end{table}

\paragraph{Practical Implementation Guidance}
Based on our comprehensive experimental evaluation, we provide concrete guidance for practitioners implementing robust NOTAM parsing systems in operational environments. The systematic comparison of prompting strategies establishes a clear hierarchy: 5-shot ICL represents the optimal approach for achieving high accuracy and reliability, while advanced reasoning techniques like CoT should be applied selectively based on specific domain requirements and model capabilities.

Domain-specific optimization emerges as a critical success factor, with careful attention to field naming conventions, output format specification, and example selection proving essential for achieving production-ready performance levels. The investigation into model scale effects provides valuable insights for deployment scenarios with varying computational constraints, enabling informed decisions about model selection and prompt optimization strategies.

Temperature analysis confirms that deterministic settings (temperature=0.0) are non-negotiable for safety-critical applications, while SRCV techniques show promise for targeted error correction but require careful validation before operational deployment. These findings establish a foundation for reliable automated processing of aviation safety communications, contributing to enhanced operational efficiency and safety in global aviation systems.

The experimental results demonstrate that while traditional information extraction methods struggle with the semantic complexity of NOTAM texts, LLM-based approaches with appropriate prompt engineering can achieve high performance across diverse operational domains. The multi-agent framework for field discovery provides additional capabilities for identifying novel, operationally critical information that might otherwise be overlooked by conventional systems.

\section{Conclusion}

This paper introduces NOTAM parsing, a novel semantic parsing task that extends beyond traditional information extraction to enable deep understanding and structured interpretation of aviation safety communications. Unlike conventional approaches focused on surface-level keyword extraction, NOTAM parsing emphasizes semantic inference, domain knowledge integration, and the generation of operationally relevant structured outputs.

Our contributions are threefold. First, we introduce Knots, a comprehensive dataset of 12,347 expert-annotated NOTAMs from 194 Flight Information Regions, whose non-extractive, inferential annotation scheme provides a robust foundation for models with genuine semantic understanding. Second, we propose a novel multi-agent collaborative framework, MDA-HDF, which systematically discovers and refines new, operationally critical information fields, achieving a 92\% F1-score in automated field discovery that significantly outperforms traditional and single-LLM methods. Third, we establish a set of optimized prompting strategies for the parsing task, demonstrating that 5-shot In-Context Learning (ICL) is the most effective approach, yielding F1-scores up to 96\%. Our analysis provides critical guidelines for practical implementation, confirming that domain-specific optimizations and deterministic temperature settings are essential for achieving safety-critical reliability.

Our work focuses on prompt engineering rather than fine-tuning. This approach establishes a strong baseline for the NOTAM parsing task by evaluating the "out-of-the-box" capabilities of LLMs, demonstrating their robust zero-shot and few-shot learning abilities within the aviation domain. This highlights the complementary nature of our contributions: the Knots dataset provides a solid foundation for future supervised fine-tuning, while our validated prompting strategies can further enhance both pre-trained and fine-tuned models.

\noindent \textbf{Limitations and Future Work:} While this work provides a foundational approach to the NOTAM parsing problem, several limitations suggest avenues for future investigation. A performance gap remains in fully resolving the task, particularly for NOTAMs involving highly complex linguistic structures or deep inferential reasoning. Additionally, the generalizability of our multi-agent framework may be constrained, as its architecture and expert-crafted prompts were specifically tailored to the semantic nuances of NOTAMs.

Future work can build upon this foundation in several promising directions. A primary avenue is to leverage the Knots dataset for supervised fine-tuning of Large Language Models (LLMs), which is expected to further enhance model performance and domain adaptation. Furthermore, expanding our methodology to other safety-critical domains could validate the generalizability of our enhanced dataset construction and prompt optimization principles. 

\appendix

\section{Theoretical Proofs and Example Prompts}

This appendix provides formal mathematical proofs of the theoretical results presented in Section~\ref{sec:framework}, followed by illustrative examples of the prompts used in our multi-agent framework.

\subsection{Proof of Lemma 1 (Conceptual Space Expansion in MDA)}

\begin{proof}
Define \(I_0 = \{t\}\) as the initial information (raw NOTAM text). Each agent's output \(Z_k\) is conditioned on all preceding context \(I_{k-1} = I_0 \cup Z_1 \cup \dots \cup Z_{k-1}\).

Because \(I_k \supseteq I_{k-1}\), the reachable latent concept space satisfies:
\[
\Theta(I_k) \supseteq \Theta(I_{k-1})
\]

Agents with complementary knowledge properties sample new regions:
\[
Z_k \sim P(\theta \mid \phi_k, I_{k-1}), \quad \theta \in \Theta(I_{k-1})
\]

Aggregating outputs through the consensus mechanism yields:
\[
Z_{\mathrm{MDA}} = \text{ConsensusAggregator}(Z_1 \cup Z_2 \cup Z_3, \tau)
\]
where the aggregator merges semantically similar fields. Since the aggregation preserves conceptual content while potentially reducing redundancy:
\[
\Theta(Z_{\mathrm{MDA}}) \supseteq \Theta(Z_1) \cup \Theta(Z_2) \cup \Theta(Z_3)
\]

Using monotonicity of \(\mu\):
\[
\mu(\Theta(Z_{\mathrm{MDA}})) \geq \max_{k \in \{1,2,3\}} \mu(\Theta(Z_k))
\]

This completes the proof.
\end{proof}

\subsection{Proof of Theorem 1 (Robustness of the HDF)}

\begin{proof}
The proof consists of two parts:

\textbf{(a) Adversarial Critique Reduces Erroneous Proposal Acceptance}

By the Misconception Refutation principle \cite{Estornell2024}, the introduction of a refuting critique \(z_{\mathrm{critique}}\) reduces the posterior probability of erroneous concepts \(\theta'\):
\[
P(\theta' \mid z_{\mathrm{proposal}}, z_{\mathrm{critique}}) < P(\theta' \mid z_{\mathrm{proposal}})
\]

Acceptance probability of a proposal correlates positively with this posterior, so:
\[
P_{\mathrm{reject}}(z_{\mathrm{proposal}} \mid z_{\mathrm{critique}}) > P_{\mathrm{reject}}(z_{\mathrm{proposal}})
\]

This shows \(A_{\mathrm{critic}}\) effectively suppresses flawed notions.

\textbf{(b) Partitioning and Deterministic Rules Mitigate Majority Tyranny}

Voting-based debates amplify the majority latent concept \(\theta'_{\mathrm{maj}}\) via peer pressure, suppressing minority voices. Conceptual partitioning into orthogonal subspaces (e.g., structural \(\Theta_S\) and terminological \(\Theta_T\)) isolates dominance in one space from the others. The deterministic FieldManager \(M_{\mathrm{field}}\) applies proposals only if unchallenged by \(A_{\mathrm{critic}}\), eliminating probabilistic "volume-based" voting and thereby removing the root cause of tyranny.

Hence,
\[
\mathcal{P}_{\mathrm{HDF}}(\mathrm{converge\ to\ } \theta'_{\mathrm{maj}}) < \mathcal{P}_{\mathrm{Vanilla}}(\mathrm{converge\ to\ } \theta'_{\mathrm{maj}})
\]

Establishing greater robustness of HDF.
\end{proof}

\subsection{HDF Workflow: A Case Study}
\label{sec:appendix_case_study}

\begin{figure}[htbp]
\centering
\includegraphics[width=0.49\textwidth]{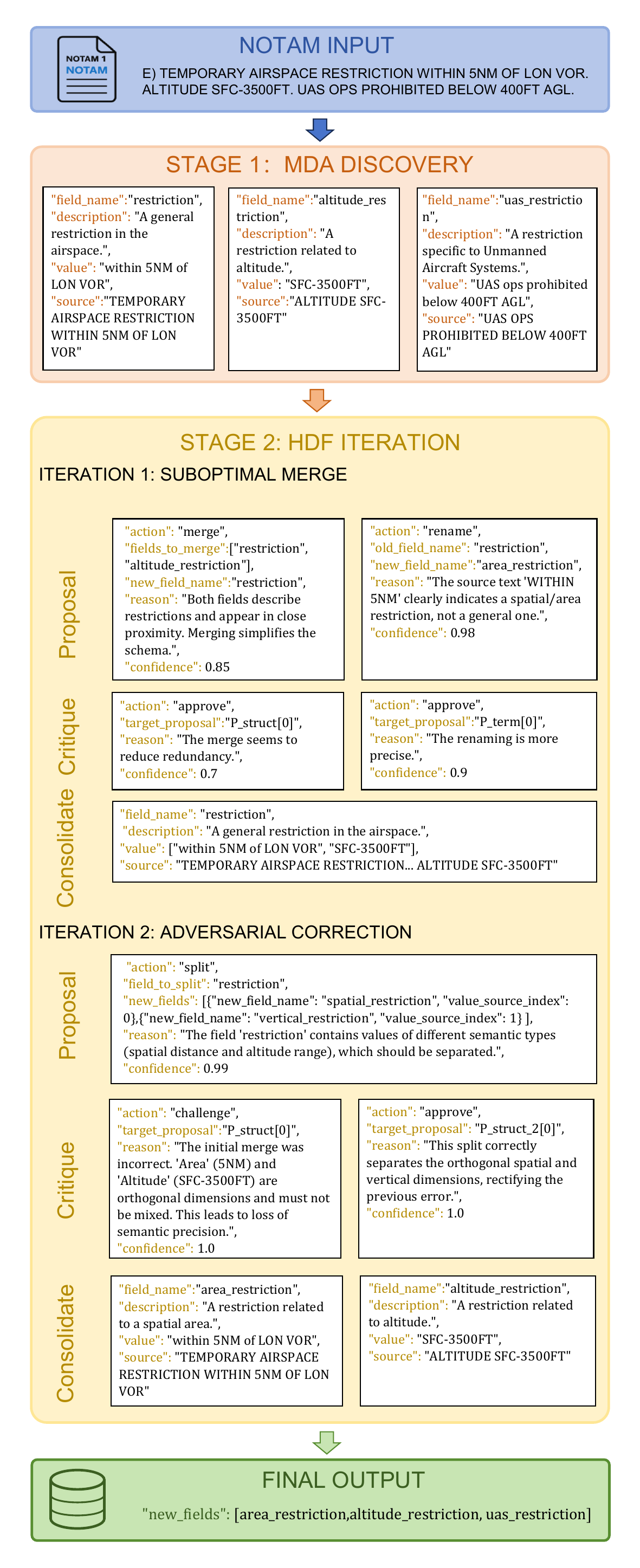}
\caption{An illustrative case study of the HDF's multi-iteration workflow. The process begins with proposals and a rule-based decision in Iteration 1. A subsequent iteration is triggered to refine the outcome based on a challenge from the Critic Agent. As defined in Algorithm~\ref{alg:mda-hdf}, this iterative cycle continues until a stable state is reached (i.e., no new proposals are generated), ensuring a robust and converged final output.}
\label{fig:example}
\end{figure}

Figure~\ref{fig:example} illustrates a complete refinement cycle of the Hybrid Debate Framework (HDF), showing how initial agent proposals, rule-based acceptance, and a critic-triggered corrective pass interact. The sequence makes explicit how iterative proposal, targeted adversarial review, and deterministic consolidation reduce redundancy and resolve ambiguity to yield a stable final specification.

\subsection{Example Prompts}

We present representative prompts used in our framework to demonstrate how semantic parsing and field refinement are systematically implemented. These examples illustrate the prompt engineering strategies that enable effective NOTAM information extraction and processing.

\textbf{NOTAM Parsing Prompt:} Demonstrates the core NOTAM semantic parsing task, showing how complex aviation terminology and domain knowledge are integrated to extract structured information from raw NOTAM text. This example focuses on runway lighting systems with detailed mapping rules and status determination logic.

\begin{promptbox}{NOTAM Runway Lighting parsing Prompt}
As an AI assistant specialized in processing NOTAM runway lighting information, extract unavailable/degraded lighting systems according to the following rules:

\textbf{Scope:} Focus only on runway lighting anomalies (RCL/REDL/RTZL/ALS). Ignore non-runway lighting (e.g., taxiway lights).

\textbf{Lighting Type Mapping:} \\
\{"REDL": ["EDGE", "REDL", "EDGE LGT"], "ALS": ["APCH", "APPROACH", "ALS", "PALS"], "RCL": ["CENTERLINE", "RCL", "CL"], "RTZL": ["TOUCHDOWN", "TDZ", "RTZL"]\}

\vspace{1ex}
If multiple systems are affected simultaneously, split into separate records.
\vspace{1ex}

\textbf{Status Determination:}
\begin{itemize}
    \item RCL/REDL/RTZL: Partial unavailability = Full unavailability (unavailable/downgrade = "unavailable")
    \item ALS Grading:
    \begin{itemize}[leftmargin=*]
        \item If status contains U/S or UNSERVICEABLE: unavailable/downgrade = "unavailable"
        \item Else: unavailable/downgrade = "downgrade"
    \end{itemize}
    \item als\_downgrade = \{"distance $\ge$ 720m": "FALS", "420-719m": "IALS", "210-419m": "BALS", "$<$210m": "NALS", "percentage/partial damage": "BALS"\}
\end{itemize}

\textbf{ILS Category Downgrade:} If "DOWNGRADED TO CAT X" appears, mark canceled ILS categories in ilscategory field (CAT-I/CAT-II/CAT-III).

\textbf{Output Format:} \\
\texttt{airport}: ICAO code \\
\texttt{runway}: runway number \\
\texttt{lightcategory}: REDL / ALS / RCL / RTZL \\
\texttt{ilscategory}: CAT-I | CAT-II | CAT-III | null \\
\texttt{unavailable/downgrade}: unavailable | downgrade \\
\texttt{als}: FALS | IALS | BALS | NALS | null \\
\texttt{distance}: distance value or null \\
\texttt{percentage}: percentage value or null

\textbf{Priority:} Explicit distance $>$ Percentage description $>$ Conservative estimation.

\vspace{1ex}

Now, based on above requirements, extract relevant information from the given NOTAM text and output in JSON.
Input:\{NOTAM\}
\end{promptbox}

\textbf{Field Consolidation Prompt:} Focuses on semantic overlap detection and field merging, employing domain knowledge to identify redundant or overlapping fields and propose consolidation strategies based on semantic similarity analysis.

\begin{promptbox}{Consolidator Prompt}
You are a specialized NOTAM field merging expert. Your task is to analyze NOTAM field definitions to identify semantically duplicate or inclusive fields.

\textbf{Focus on:}
\begin{itemize}
    \item \textbf{Semantic overlap of field names}: e.g., \texttt{runway\_closure} vs \texttt{runway\_closed}
    \item \textbf{Duplicate content in descriptions}: e.g., \texttt{ice} vs \texttt{ice\_condition}
    \item \textbf{High overlap in source content}: same NOTAM patterns or identical meanings
\end{itemize}

\textbf{Output Requirements:}
\begin{itemize}
    \item Output only a JSON array
    \item Each object must contain: \texttt{"action"}, \texttt{"fields\_to\_merge"}, \texttt{"new\_field\_name"}, \texttt{"reason"}, \texttt{"confidence"}
    \item Confidence is a float value from 0.0 to 1.0
\end{itemize}

\textbf{Example:} \\
\texttt{[ \{ "action": "merge", "fields\_to\_merge": ["runway\_closure", "runway\_closed"], "new\_field\_name": "runway\_closure", "reason": "Both fields indicate 'RWY CLSD' and share identical semantics.", "confidence": 0.95 \} ]} \\[1ex]

If no fields require merging, return: \texttt{[]}
\end{promptbox}

\textbf{Terminology Standardization Prompt:} Responsible for renaming fields using precise aviation terminology, ensuring that discovered fields conform to industry standards and best practices through systematic terminology validation and standardization.

\begin{promptbox}{Specializer Prompt}
You are an aviation terminology expert. Analyze NOTAM field definitions and, if a field name is too generic but its sources contain more precise terms, suggest renaming it.

\textbf{Focus on:}
\begin{itemize}
    \item Check for professional terms in sources
    \item Suggest industry-standard terminology
    \item Provide clear renaming suggestions
\end{itemize}

\textbf{Output Requirements:}
\begin{itemize}
    \item Output only a JSON array
    \item Each object must contain: \texttt{"action"}, \texttt{"old\_field\_name"}, \texttt{"new\_field\_name"}, \texttt{"reason"}, \texttt{"confidence"}
    \item Confidence is a float value from 0.0 to 1.0
\end{itemize}

\textbf{Example:} \\
\texttt{[ \{ "action": "rename", "old\_field\_name": "declared\_distance", "new\_field\_name": "TORA", "reason": "The sources explicitly contain 'TORA 8102FT'; precise terminology should be used.", "confidence": 1.0 \} ]} \\[1ex]

If no fields require renaming, return: \texttt{[]}
\end{promptbox}

\textbf{Quality Control Prompt:} Provides evaluation and quality control for merging and renaming proposals, implementing adversarial critique mechanisms to challenge potentially flawed suggestions and ensure robustness through systematic review processes.

\begin{promptbox}{Critic Prompt}
You are a critical analysis expert. Review previous merging and renaming proposals and identify potential issues.

\textbf{Focus on:}
\begin{itemize}
    \item Check if merge proposals are too aggressive and could cause information loss
    \item Verify that renaming proposals are accurate and retain important context
    \item Identify any duplicate fields that were missed
\end{itemize}

\textbf{Output Requirements:}
\begin{itemize}
    \item Output only a JSON array
    \item Each object must contain: \texttt{"action"}, \texttt{"target\_proposal"}, \texttt{"reason"}, \texttt{"confidence"}
    \item \texttt{"action"} is either \texttt{"challenge"} or \texttt{"approve"}
    \item Confidence is a float value from 0.0 to 1.0
\end{itemize}

\textbf{Example:} \\
\texttt{[ \{ "action": "challenge", "target\_proposal": 0, "reason": "Although 'hazard' and 'hazard\_area' are similar, 'hazard\_area' emphasizes spatial scope and should not be merged.", "confidence": 0.8 \} ]} \\[1ex]

If all proposals are fine, return: \texttt{[]}
\end{promptbox}

These prompt examples demonstrate the systematic approach to NOTAM parsing and field refinement, where specialized prompts handle different aspects of the semantic processing pipeline while maintaining consistency and domain expertise.

\bibliographystyle{elsarticle-num} 
\bibliography{main}

\begin{thebibliography}{10}
\expandafter\ifx\csname url\endcsname\relax
  \def\url#1{\texttt{#1}}\fi
\expandafter\ifx\csname urlprefix\endcsname\relax\def\urlprefix{URL }\fi
\expandafter\ifx\csname href\endcsname\relax
  \def\href#1#2{#2} \def\path#1{#1}\fi

\bibitem{arnold2022knowledge}
A.~Arnold, F.~Ernez, C.~Kobus, M.-C. Martin, Knowledge extraction from aeronautical messages (notams) with self-supervised language models for aircraft pilots, in: Proceedings of NAACL-HLT 2022: Industry Track Papers, 2022, pp. 188--196.

\bibitem{bravin2020automated}
M.~Bravin, S.~Mazumder, D.~Pfäffli, M.~Pouly, Automated smartification of notices to airmen, in: Proceedings of the 7th Swiss Conference on Data Science (SDS), 2020, pp. 51--52.

\bibitem{clarke2021natural}
S.~S.~B. Clarke, P.~Maynard, J.~A. Almache, S.~G. Kumar, S.~Rajkumar, A.~C. Kemp, R.~Pai, Natural language processing analysis of notices to airmen for air traffic management optimization, in: Proceedings of the AIAA Aviation Forum 2021, 2021, pp. 1--26.

\bibitem{morarasu2024aidriven}
M.~M. Morarasu, C.~H. Roman, Ai-driven optimization of operational notam management, in: 2024 Integrated Communications, Navigation and Surveillance Conference (ICNS), IEEE, 2024, pp. 1--6.
\newblock \href {https://doi.org/10.1109/ICNS60906.2024.10550725} {\path{doi:10.1109/ICNS60906.2024.10550725}}.

\bibitem{mogillo-dettwilerfiltering}
A.~Mogillo-Dettwiler, Filtering and sorting of notices to air missions (notams) (2024).

\bibitem{mi2022notam}
B.~Mi, Y.~Fan, Y.~Sun, Notam text analysis and classification based on attention mechanism, Journal of Physics: Conference Series 2171~(1) (2022) 012042.
\newblock \href {https://doi.org/10.1088/1742-6596/2171/1/012042} {\path{doi:10.1088/1742-6596/2171/1/012042}}.

\bibitem{guo2024integrating}
D.~Guo, Z.~Zhang, B.~Yang, J.~Zhang, H.~Yang, Y.~Lin, Integrating spoken instructions into flight trajectory prediction to optimize automation in air traffic control, Nature Communications 15~(1) (2024) 9662.

\bibitem{fan2024global}
Y.~Fan, Y.~Tan, L.~Wu, H.~Ye, Z.~Lyu, \href{https://doi.org/10.1109/TAES.2024.3357668}{Global and local interattribute relationships-based graph convolutional network for flight trajectory prediction}, {IEEE} Trans. Aerosp. Electron. Syst. 60~(3) (2024) 2642--2657.
\newblock \href {https://doi.org/10.1109/TAES.2024.3357668} {\path{doi:10.1109/TAES.2024.3357668}}.
\newline\urlprefix\url{https://doi.org/10.1109/TAES.2024.3357668}

\bibitem{luo2025large}
K.~Luo, J.~Zhou, \href{https://doi.org/10.48550/arXiv.2501.17459}{Large language models for single-step and multi-step flight trajectory prediction}, CoRR abs/2501.17459 (2025).
\newblock \href {http://arxiv.org/abs/2501.17459} {\path{arXiv:2501.17459}}, \href {https://doi.org/10.48550/ARXIV.2501.17459} {\path{doi:10.48550/ARXIV.2501.17459}}.
\newline\urlprefix\url{https://doi.org/10.48550/arXiv.2501.17459}

\bibitem{liu2024nlp}
X.~Liu, B.~Zou, A.~Aw, \href{https://doi.org/10.18653/v1/2024.naacl-industry.8}{An nlp-focused pilot training agent for safe and efficient aviation communication}, in: Y.~Yang, A.~Davani, A.~Sil, A.~Kumar (Eds.), Proceedings of the 2024 Conference of the North American Chapter of the Association for Computational Linguistics: Human Language Technologies: Industry Track, {NAACL} 2024, Mexico City, Mexico, June 16-21, 2024, Association for Computational Linguistics, 2024, pp. 89--96.
\newblock \href {https://doi.org/10.18653/V1/2024.NAACL-INDUSTRY.8} {\path{doi:10.18653/V1/2024.NAACL-INDUSTRY.8}}.
\newline\urlprefix\url{https://doi.org/10.18653/v1/2024.naacl-industry.8}

\bibitem{nanyonga2023aviation}
A.~Nanyonga, H.~Wasswa, G.~Wild, Aviation safety enhancement via nlp \& deep learning: Classifying flight phases in atsb safety reports, in: 2023 Global Conference on Information Technologies and Communications (GCITC), IEEE, 2023, pp. 1--5.

\bibitem{xu2024survey}
D.~Xu, W.~Chen, W.~Peng, C.~Zhang, T.~Xu, X.~Zhao, X.~Wu, Y.~Zheng, E.~Chen, \href{https://api.semanticscholar.org/CorpusID:266690657}{Large language models for generative information extraction: A survey}, Frontiers Comput. Sci. 18 (2023) 186357.
\newline\urlprefix\url{https://api.semanticscholar.org/CorpusID:266690657}

\bibitem{brown2020languagemodelsfewshotlearners}
T.~B. Brown, B.~Mann, N.~Ryder, M.~Subbiah, J.~Kaplan, P.~Dhariwal, et~al., \href{https://arxiv.org/abs/2005.14165}{Language models are few-shot learners} (2020).
\newblock \href {http://arxiv.org/abs/2005.14165} {\path{arXiv:2005.14165}}.
\newline\urlprefix\url{https://arxiv.org/abs/2005.14165}

\bibitem{Brown2020}
T.~B. Brown, B.~Mann, N.~Ryder, M.~Subbiah, J.~Kaplan, P.~Dhariwal, et~al., \href{https://arxiv.org/abs/2005.14165}{Language models are few-shot learners} (2020).
\newblock \href {http://arxiv.org/abs/2005.14165} {\path{arXiv:2005.14165}}.
\newline\urlprefix\url{https://arxiv.org/abs/2005.14165}

\bibitem{Sahoo2025}
P.~Sahoo, A.~K. Singh, S.~Saha, V.~Jain, S.~Mondal, A.~Chadha, A systematic survey of prompt engineering in large language models: Techniques and applications, arXiv preprint arXiv:2402.07927 (2024).

\bibitem{Wei2022}
J.~Wei, X.~Wang, D.~Schuurmans, M.~Bosma, F.~Xia, E.~Chi, Q.~V. Le, D.~Zhou, et~al., Chain-of-thought prompting elicits reasoning in large language models, Advances in neural information processing systems 35 (2022) 24824--24837.

\bibitem{Zhang2022}
Z.~Zhang, A.~Zhang, M.~Li, A.~Smola, Automatic chain of thought prompting in large language models, arXiv preprint arXiv:2210.03493 (2022).

\bibitem{Wang2022}
X.~Wang, J.~Wei, D.~Schuurmans, Q.~Le, E.~Chi, S.~Narang, A.~Chowdhery, D.~Zhou, Self-consistency improves chain of thought reasoning in language models, arXiv preprint arXiv:2203.11171 (2022).

\bibitem{zhao2023enhancing}
X.~Zhao, M.~Li, W.~Lu, C.~Weber, J.~H. Lee, K.~Chu, S.~Wermter, Enhancing zero-shot chain-of-thought reasoning in large language models through logic, arXiv preprint arXiv:2309.13339 (2023).

\bibitem{Yao2023a}
S.~Yao, D.~Yu, J.~Zhao, I.~Shafran, T.~Griffiths, Y.~Cao, K.~Narasimhan, Tree of thoughts: Deliberate problem solving with large language models, Advances in neural information processing systems 36 (2023) 11809--11822.

\bibitem{Yao2023b}
Y.~Yao, Z.~Li, H.~Zhao, Beyond chain-of-thought, effective graph-of-thought reasoning in language models, arXiv preprint arXiv:2305.16582 (2023).

\bibitem{Lewis2020}
P.~Lewis, E.~Perez, A.~Piktus, F.~Petroni, V.~Karpukhin, N.~Goyal, H.~K{\"u}ttler, M.~Lewis, W.-t. Yih, T.~Rockt{\"a}schel, et~al., Retrieval-augmented generation for knowledge-intensive nlp tasks, Advances in neural information processing systems 33 (2020) 9459--9474.

\bibitem{Dhuliawala2023}
S.~Dhuliawala, M.~Komeili, J.~Xu, R.~Raileanu, X.~Li, A.~Celikyilmaz, J.~Weston, Chain-of-verification reduces hallucination in large language models, arXiv preprint arXiv:2309.11495 (2023).

\bibitem{Yuan2024}
X.~Yuan, C.~Shen, S.~Yan, X.~Zhang, L.~Xie, W.~Wang, R.~Guan, Y.~Wang, J.~Ye, Instance-adaptive zero-shot chain-of-thought prompting, Advances in Neural Information Processing Systems 37 (2024) 125469--125486.

\bibitem{li2023codeie}
P.~Li, T.~Sun, Q.~Tang, H.~Yan, Y.~Wu, X.~Huang, X.~Qiu, {CodeIE}: Large code generation models are better few-shot information extractors, in: Proceedings of the 61st Annual Meeting of the Association for Computational Linguistics (Volume 1: Long Papers), Association for Computational Linguistics, Toronto, Canada, 2023, pp. 15339--15353.

\bibitem{wang2023instructuie}
X.~Wang, W.~Zhou, C.~Zu, H.~Xia, T.~Chen, Y.~Zhang, R.~Zheng, J.~Ye, Q.~Zhang, T.~Gui, et~al., {InstructUIE}: Multitask instruction tuning for unified information extraction, arXiv preprint arXiv:2304.08085 (2023).

\bibitem{sainz2024gollie}
O.~Sainz, I.~Garc{\'\i}a-Ferrero, R.~Agerri, O.~Lopez~de Lacalle, G.~Rigau, E.~Agirre, \href{https://openreview.net/forum?id=Y3wpuxd7u9}{Go{LLIE}: Annotation guidelines improve zero-shot information-extraction}, in: The Twelfth International Conference on Learning Representations, 2024.
\newline\urlprefix\url{https://openreview.net/forum?id=Y3wpuxd7u9}

\bibitem{li2024knowcoder}
Z.~Li, Y.~Zeng, Y.~Zuo, W.~Ren, W.~Liu, M.~Su, Y.~Guo, Y.~Liu, et~al., {KnowCoder}: Coding structured knowledge into {LLMs} for universal information extraction, in: Proceedings of the 62nd Annual Meeting of the Association for Computational Linguistics (Volume 1: Long Papers), Association for Computational Linguistics, Bangkok, Thailand, 2024, pp. 8758--8779.

\bibitem{Tran2025MultiAgentCM}
K.-T. Tran, D.~Dao, M.-D. Nguyen, Q.-V. Pham, B.~O’Sullivan, H.~D. Nguyen, \href{https://api.semanticscholar.org/CorpusID:275471465}{Multi-agent collaboration mechanisms: A survey of llms}, ArXiv abs/2501.06322 (2025).
\newline\urlprefix\url{https://api.semanticscholar.org/CorpusID:275471465}

\bibitem{xi2023}
Z.~Xi, W.~Chen, X.~Guo, W.~He, Y.~Ding, B.~Hong, M.~Zhang, J.~Wang, S.~Jin, E.~Zhou, et~al., The rise and potential of large language model based agents: A survey, Science China Information Sciences 68~(2) (2025) 121101.

\bibitem{guo2024multiagent}
T.~Guo, X.~Chen, Y.~Wang, R.~Chang, S.~Pei, N.~V. Chawla, O.~Wiest, X.~Zhang, Large language model based multi-agents: A survey of progress and challenges, arXiv preprint arXiv:2402.01680 (2024).

\bibitem{lu2024}
J.~Lu, Z.~Pang, M.~Xiao, Y.~Zhu, R.~Xia, J.~Zhang, Merge, ensemble, and cooperate! a survey on collaborative strategies in the era of large language models, arXiv preprint arXiv:2407.06089 (2024).

\bibitem{han2024}
S.~Han, Q.~Zhang, Y.~Yao, W.~Jin, Z.~Xu, Llm multi-agent systems: Challenges and open problems, arXiv preprint arXiv:2402.03578 (2024).

\bibitem{liang2023}
T.~Liang, Z.~He, W.~Jiao, X.~Wang, Y.~Wang, R.~Wang, Y.~Yang, S.~Shi, Z.~Tu, Encouraging divergent thinking in large language models through multi-agent debate, arXiv preprint arXiv:2305.19118 (2023).

\bibitem{zhang2023}
J.~Zhang, X.~Xu, N.~Zhang, R.~Liu, B.~Hooi, S.~Deng, Exploring collaboration mechanisms for llm agents: A social psychology view, arXiv preprint arXiv:2310.02124 (2023).

\bibitem{li2023a}
H.~Li, Y.~Q. Chong, S.~Stepputtis, J.~Campbell, D.~Hughes, M.~Lewis, K.~Sycara, Theory of mind for multi-agent collaboration via large language models, arXiv preprint arXiv:2310.10701 (2023).

\bibitem{li2023b}
G.~Li, H.~Hammoud, H.~Itani, D.~Khizbullin, B.~Ghanem, Camel: Communicative agents for" mind" exploration of large language model society, Advances in Neural Information Processing Systems 36 (2023) 51991--52008.

\bibitem{wu2024}
Q.~Wu, G.~Bansal, J.~Zhang, Y.~Wu, B.~Li, E.~Zhu, L.~Jiang, X.~Zhang, S.~Zhang, J.~Liu, et~al., Autogen: Enabling next-gen llm applications via multi-agent conversations, in: First Conference on Language Modeling, 2024.

\bibitem{he2025}
J.~He, C.~Treude, D.~Lo, Llm-based multi-agent systems for software engineering: Literature review, vision, and the road ahead, ACM Transactions on Software Engineering and Methodology 34~(5) (2025) 1--30.

\bibitem{lambiase2024}
S.~Lambiase, G.~Catolino, F.~Palomba, F.~Ferrucci, Motivations, challenges, best practices, and benefits for bots and conversational agents in software engineering: A multivocal literature review, ACM Computing Surveys 57~(4) (2024) 1--37.

\bibitem{Estornell2024}
A.~Estornell, Y.~Liu, Multi-llm debate: Framework, principals, and interventions, in: A.~Globerson, L.~Mackey, D.~Belgrave, A.~Fan, U.~Paquet, J.~Tomczak, C.~Zhang (Eds.), Advances in Neural Information Processing Systems, Vol.~37, Curran Associates, Inc., 2024, pp. 28938--28964.

\bibitem{Huang2023ASO}
L.~Huang, W.~Yu, W.~Ma, W.~Zhong, Z.~Feng, H.~Wang, Q.~Chen, W.~Peng, X.~Feng, B.~Qin, T.~Liu, A survey on hallucination in large language models: Principles, taxonomy, challenges, and open questions, ACM Transactions on Information Systems 43 (2023) 1 -- 55.

\bibitem{zhang2025iopo}
X.~Zhang, H.~Yu, C.~Fu, F.~Huang, Y.~Li, \href{https://aclanthology.org/2025.acl-long.1079/}{{IOPO}: Empowering {LLM}s with complex instruction following via input-output preference optimization}, in: Proceedings of the 63rd Annual Meeting of the Association for Computational Linguistics (Volume 1: Long Papers), Association for Computational Linguistics, 2025.
\newline\urlprefix\url{https://aclanthology.org/2025.acl-long.1079/}

\end{thebibliography}

\end{document}